\RequirePackage{fix-cm}
\documentclass[10pt,journal,compsoc]{IEEEtran}
\usepackage[numbers,sort&compress]{natbib}
\usepackage[utf8]{inputenc}
\usepackage[T1]{fontenc}
\usepackage{url}
\usepackage{booktabs}
\usepackage{amsfonts}
\usepackage{amsmath}
\usepackage{amsthm}
\usepackage{algorithm}
\usepackage{algpseudocode}
\newtheorem{lemma}{Lemma}
\newtheorem{proposition}{Proposition}
\newtheorem{corollary}{Corollary}

\usepackage{enumitem}
\usepackage{nicefrac}
\usepackage{microtype}
\usepackage{ragged2e}
\usepackage{xcolor}
\usepackage{colortbl}
\usepackage{listings}
\definecolor{codebg}{HTML}{F7F7F8}
\definecolor{codegreen}{HTML}{1B5E20}
\definecolor{codepurple}{HTML}{4A148C}
\definecolor{codegray}{HTML}{757575}
\lstdefinestyle{pythonstyle}{
  language=Python,
  backgroundcolor=\color{codebg},
  basicstyle=\ttfamily\scriptsize,
  keywordstyle=\color{codepurple}\bfseries,
  commentstyle=\color{codegray}\itshape,
  stringstyle=\color{codegreen},
  numbers=left,
  numberstyle=\tiny\color{codegray},
  numbersep=6pt,
  frame=single,
  rulecolor=\color{codegray!30},
  breaklines=true,
  showstringspaces=false,
  tabsize=4,
  xleftmargin=1.5em,
  framexleftmargin=1em,
}
\usepackage{graphicx}
\usepackage{pgfplots}
\usetikzlibrary{calc}
\usetikzlibrary{arrows.meta,positioning,fit,shapes.geometric,backgrounds}
\usepgfplotslibrary{groupplots}
\pgfplotsset{compat=1.18}
\usepackage{float}
\usepackage{placeins}
\usepackage{afterpage}
\usepackage[most]{tcolorbox}
\usepackage{pifont}
\usepackage{fontawesome5}
\definecolor{mydarkblue}{HTML}{1A4F8B}
\newsavebox{\tablefitbox}
\newcommand{\fitwidth}[1]{%
  \sbox{\tablefitbox}{#1}%
  \ifdim\wd\tablefitbox>\linewidth
    \resizebox{\linewidth}{!}{\usebox{\tablefitbox}}%
  \else
    \makebox[\linewidth][c]{\usebox{\tablefitbox}}%
  \fi
}
\usepackage{hyperref}
\hypersetup{
  colorlinks=true,
  linkcolor=mydarkblue,
  citecolor=mydarkblue,
  urlcolor=mydarkblue,
  filecolor=mydarkblue
}
\newtcolorbox{problembox}{
  enhanced,
  colback=mydarkblue!4,
  colframe=mydarkblue!22,
  boxrule=0.4pt,
  arc=1.2mm,
  left=1.2mm,
  right=1.2mm,
  top=0.9mm,
  bottom=0.9mm,
  borderline west={1.4pt}{0pt}{mydarkblue!55}
}

\newtcolorbox{contribbox}{
  enhanced,
  breakable,
  colback=mydarkblue!2,
  colframe=mydarkblue!24,
  boxrule=0.4pt,
  arc=1.1mm,
  left=1.5mm,
  right=1.5mm,
  top=1.0mm,
  bottom=0.9mm,
  borderline west={1.6pt}{0pt}{mydarkblue!60}
}

\DeclareRobustCommand{\roboimagine}{\textrm{\textsc{RoboImagine}}}
\definecolor{tblheadbg}{HTML}{E8EBF3}
\definecolor{tblstripe}{HTML}{F7F8FB}
\definecolor{rankfirstbg}{HTML}{FCE9E7}
\definecolor{ranksecondbg}{HTML}{FEF1E5}
\definecolor{rankthirdbg}{HTML}{E8F7FB}
\newcommand{\scorestd}[2]{#1{\scriptsize$_{\pm#2}$}}
\newcommand{\bestscore}[2]{\cellcolor{rankfirstbg}\textbf{\scorestd{#1}{#2}}\textsuperscript{\textcolor{red!70!black}{\scriptsize 1}}}
\newcommand{\secondscore}[2]{\cellcolor{ranksecondbg}\textbf{\scorestd{#1}{#2}}\textsuperscript{\textcolor{orange!85!black}{\scriptsize 2}}}
\newcommand{\thirdscore}[2]{\cellcolor{rankthirdbg}\textbf{\scorestd{#1}{#2}}\textsuperscript{\textcolor{cyan!60!black}{\scriptsize 3}}}
\newcommand{\plainscore}[2]{\scorestd{#1}{#2}}
\newcommand{\bestmean}[1]{\cellcolor{rankfirstbg}\textbf{#1}\textsuperscript{\textcolor{red!70!black}{\scriptsize 1}}}
\newcommand{\secondmean}[1]{\cellcolor{ranksecondbg}\textbf{#1}\textsuperscript{\textcolor{orange!85!black}{\scriptsize 2}}}
\newcommand{\thirdmean}[1]{\cellcolor{rankthirdbg}\textbf{#1}\textsuperscript{\textcolor{cyan!60!black}{\scriptsize 3}}}

\newsavebox{\maincomparisonbox}
\NewDocumentEnvironment{fullwidthcomparison}{m m +b}{%
  \sbox{\maincomparisonbox}{\begin{tabular}{#2}#3\end{tabular}}%
  \setlength{\tabcolsep}{\dimexpr\tabcolsep+(\textwidth-\wd\maincomparisonbox)/#1/2\relax}%
  \begin{tabular*}{\textwidth}{#2}#3\end{tabular*}%
}{}

\hypersetup{pdftitle={Learning Robotic World Models with Semantic and Visual Feedback},pdfauthor={Hao Wu; Yuqi Li; Yizhou Zhao; Kun Wang; Xiao Luo; Yu Cheng; Bo An; Yingli Tian; Laszlo A. Jeni; Ming-Hsuan Yang},pdfsubject={Robotic world models - arXiv preprint}}

\title{Learning Robotic World Models\\with Semantic and Visual Feedback}
\author{Hao Wu, Yuqi Li, Yizhou Zhao, Kun Wang, Xiao Luo, Yu Cheng, Bo An,\\
Yingli Tian,~\IEEEmembership{Fellow,~IEEE}, Laszlo A. Jeni, and Ming-Hsuan Yang\textsuperscript{\faEnvelope[regular]},~\IEEEmembership{Fellow,~IEEE}%
\IEEEcompsocitemizethanks{%
\IEEEcompsocthanksitem Hao Wu, Yizhou Zhao, and Laszlo A. Jeni are with CMU (e-mail:
\href{mailto:hw5@andrew.cmu.edu}{hw5@andrew.cmu.edu}).
\IEEEcompsocthanksitem Yuqi Li and Yingli Tian are with The City College of New York.
\IEEEcompsocthanksitem Kun Wang, Yu Cheng, and Bo An are with NTU.
\IEEEcompsocthanksitem Xiao Luo is with the University of Wisconsin--Madison.
\IEEEcompsocthanksitem Ming-Hsuan Yang is with UC Merced. \textsuperscript{\faEnvelope[regular]}Corresponding author: Ming-Hsuan Yang.}}

\IEEEtitleabstractindextext{%
\vspace{-8pt}
\par
{\centering\normalfont\small
\href{https://github.com/Alexander-wu/RoboImagine}{\textcolor{black}{\faGithub}\enspace\textcolor[HTML]{FB3199}{\textbf{Project \& Code}}}
\qquad
\href{https://huggingface.co/easylearning/RoboImagine}{\raisebox{-1.5pt}{\includegraphics[height=10pt]{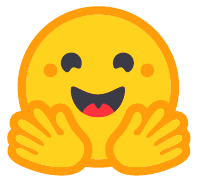}}\enspace\textcolor[HTML]{FB3199}{\textbf{Data \& Weights (Hugging Face)}}}\par}
\vspace{5pt}
\begin{abstract}
\justifying
Robot world models predict action-conditioned future visual states to support robotic planning and decision making. However, visual reconstruction does not fully capture task progress, semantic evaluation alone provides limited constraints on fine-grained visual errors, and homogeneous predictions limit comparative feedback. We present \roboimagine{}, a framework for post-training autoregressive robotic world models through candidate-future comparisons with semantic and visual feedback. Diverse search promotes diversity among candidate trajectories under shared visual observations, action sequences, and task instructions. A frozen multimodal evaluator provides task-relevant semantic preferences through pairwise video comparisons in both presentation orders, complemented by pixel- and perceptual-level visual consistency feedback that constrains local structural and appearance errors. These signals drive policy optimization through group-relative advantages. Fixed initial context and branch-specific rolling histories support long-sequence generation with bounded context and consistent conditioning between candidate generation and policy optimization. Experiments with separately trained models on RT-1, BridgeV2, and RoboCasa cover real-world and simulated manipulation. On RT-1, under matched initialization, update counts, and evaluation protocols, \roboimagine{} improves PSNR by 1.26\,dB and reduces LPIPS by 11.93\% relative to continued supervised fine-tuning. Ablations show that diverse search improves prediction quality and that combined semantic and visual feedback yields higher external multimodal evaluation scores than visual feedback alone while maintaining comparable visual fidelity. Rolling histories further improve long-horizon prediction. These results support improving visual fidelity, task relevance, and long-horizon prediction through candidate comparison and complementary feedback.

\end{abstract}
\begin{IEEEkeywords}
Robotic world models, reinforcement learning, preference feedback,
multimodal judges, visual consistency, diverse search.
\end{IEEEkeywords}}

\IEEEaftertitletext{\vspace{-22pt}}

\begin{document}
\maketitle
\IEEEdisplaynontitleabstractindextext
\IEEEpeerreviewmaketitle

\IEEEraisesectionheading{\section{Introduction}\label{sec:introduction}}
\begin{figure}[!t]
  \vspace{-2pt}
  \centering
  \includegraphics[width=\columnwidth]{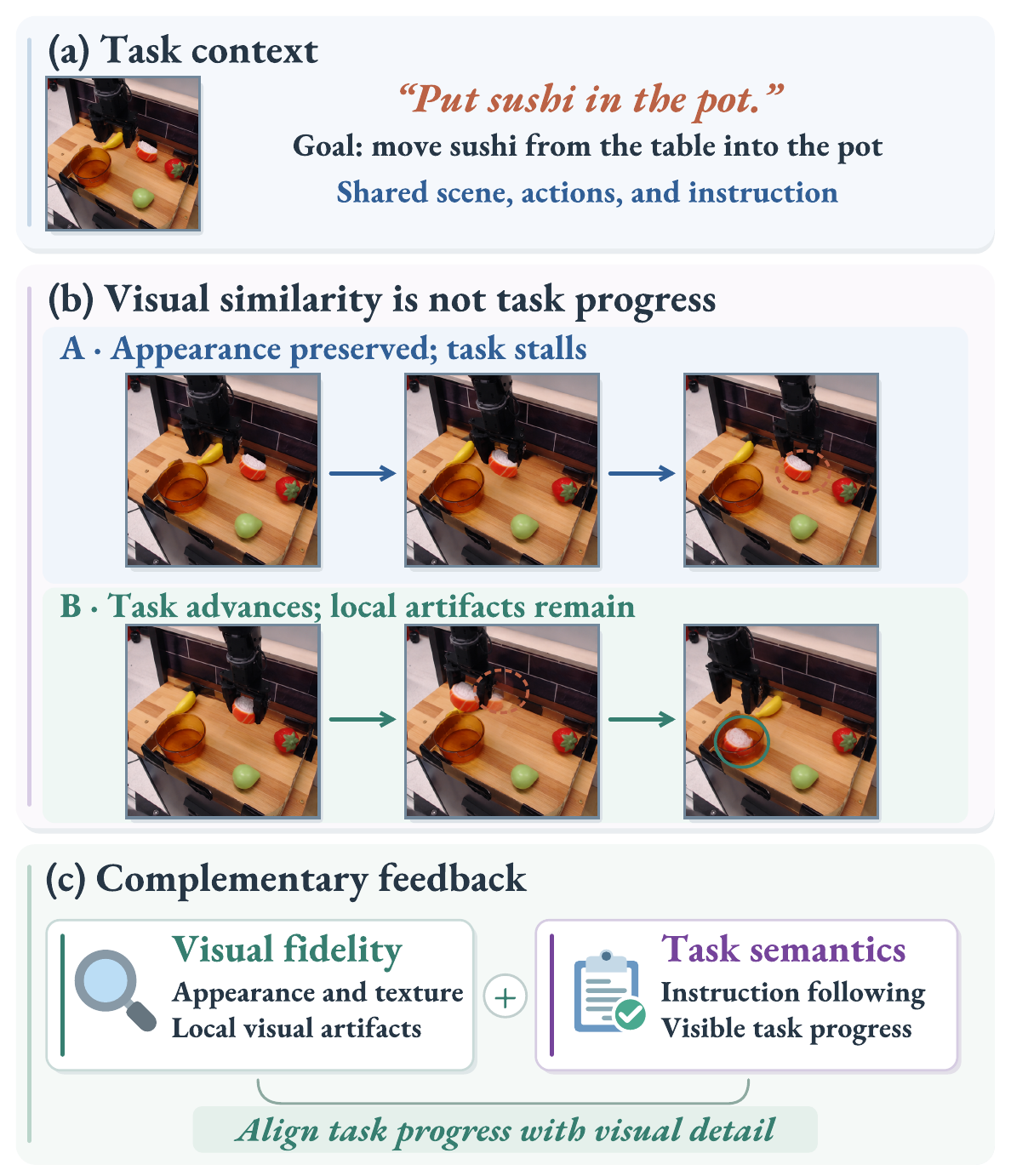}
  \vspace{-8pt}
  \caption{\textbf{Complementary feedback for robot world models.} Preserved appearance can coexist with stalled task progress, while task advancement can coexist with local visual artifacts. Visual fidelity and task semantics therefore provide complementary feedback. Scenes are conceptual illustrations, not experimental predictions or measured rankings.}
  \label{fig:motivation_feedback}
  \vspace{-30pt}
\end{figure}

Robot world models predict how scenes evolve under actions, enabling robots to assess potential outcomes before execution~\cite{ha2018world,ni2025maskgwm,brooks2024video,hou2026worldmodel}. As learnable internal simulators, they support planning, policy evaluation, and long-horizon reasoning~\cite{tan2026towards,hansen2023td,li2024evaluating,zhang2026thinkingvideo}. Unlike generic video generation, which primarily emphasizes visual quality, robot world models must capture action-induced state changes, object contacts, and interactions to provide reliable guidance for manipulation decisions~\cite{li2025worldmodelbench,wu2025rlvr}. Visual observations describe the current scene, action sequences specify the execution process, and natural-language instructions provide task goals; together, they form the multimodal context for understanding and predicting robot interactions. Prediction quality therefore depends not only on visual fidelity, but also on how accurately the model captures physical evolution under the given actions and its relationship to the task goal~\cite{barcellona2024dream,pang2025learning}.

However, evaluating predicted futures remains a central challenge in training robot world models. Existing methods typically use maximum-likelihood, reconstruction, or perceptual objectives~\cite{wu2025rlvr,huang2025vid2world,mao2025robot,ding2025understanding}. These objectives provide effective visual supervision, but visual similarity alone does not fully capture task-relevant changes. For example, a prediction that preserves scene appearance may still miss action-induced object displacement or changes in contact. Conversely, multimodal semantic evaluation assesses visible task progress using instructions and scene context, yet may be insufficiently sensitive to local structural distortions and fine-grained appearance errors. As Figure~\ref{fig:motivation_feedback} illustrates, visual fidelity and task semantics provide complementary perspectives on prediction quality. Aligning robot world models therefore calls for both forms of feedback: constraining visual detail while assessing whether predicted state transitions are plausible under the given actions and task conditions.

Prior work explores reward-based post-training for robotic world models~\cite{wu2026futureworlds}. RLVR-World rewards action-conditioned forecasts using visual prediction metrics~\cite{wu2025rlvr}. RoboAlign-R1~\cite{wu2026roboalignr1} distills a vision--language judge into a lightweight reward model that scores individual generated videos. Such semantic feedback motivates a broader question: how should alternative futures be constructed and evaluated to provide informative learning signals? Absolute video scores require calibration across scenes and tasks, whereas futures generated under shared observations, actions, and instructions can be compared directly for visible task progress. However, nearly identical candidates provide little basis for comparison, and semantic preferences alone may overlook local visual errors. Effective comparative learning must therefore address both candidate construction and feedback design. \textbf{\textit{We investigate how diverse candidate futures and complementary semantic and visual feedback jointly shape the learning of robotic world models.}}

We propose \textbf{\roboimagine{}}, a framework for learning autoregressive robotic world models by comparing candidate futures with semantic and visual feedback. Conditioned on visual observations, actions, and task instructions, grouped diverse beam search (GDBS)~\cite{vijayakumar2016diverse} constructs alternative trajectories under shared inputs. A frozen multimodal evaluator, mJev (hereafter Jev)~\cite{somark2026mjev}, compares candidate videos for visible task progress using the instruction and initial observation; evaluating both presentation orders mitigates position bias. Pixel- and perceptual-level visual consistency feedback~\cite{zhang2018unreasonable} complements these preferences by constraining fine-grained prediction errors. The two signals are standardized across candidates and combined into group-relative advantages for policy optimization~\cite{shao2024deepseekmath}. Each branch retains fixed initial context and a dynamic anchor together with its own rolling state--action history. This bounded history supports continued generation, while replaying the recorded history keeps policy scoring consistent with the conditions under which each candidate is generated.

We evaluate separately trained models on RT-1, BridgeV2, and RoboCasa, and conduct controlled studies of post-training, candidate search, feedback composition, and memory. On RT-1, comparative post-training improves PSNR by 1.26~dB and reduces LPIPS by 11.93\% relative to continued supervised fine-tuning from the same initialization for the same number of updates, at a higher computational cost. Visual-only feedback achieves the strongest visual metrics in the reward ablation, whereas combined feedback yields a higher external multimodal evaluation mean while retaining comparable visual fidelity. Long-horizon evaluations further assess the benefits of rolling history and the associated inference costs. Our main contributions are summarized as follows:

\textbf{\ding{182} \textit{Learning robot world models by comparing imagined futures}.} We introduce \roboimagine{}, which couples GDBS with comparative post-training under shared observations, actions, and instructions. Bounded branch-specific histories support autoregressive rollouts and are replayed during policy scoring to preserve conditioning consistency.

\textbf{\ding{183} \textit{Complementary semantic preference and visual consistency feedback}.} A frozen multimodal judge evaluates pairwise task progress among candidates with bidirectional comparisons to mitigate order bias, while pixel- and perceptual-level consistency feedback constrains fine-grained visual errors. Group-wise standardization and relative advantages translate both signals into policy updates without backpropagating through the evaluator or the video decoder.

\textbf{\ding{184} \textit{Evaluation of prediction quality and learning mechanisms}.} Experiments on three manipulation datasets assess visual prediction and task-level consistency. Controlled RT-1 comparisons isolate the effects of additional supervised training, feedback composition, and candidate search, while long-horizon and window-size studies characterize the quality and inference cost of bounded memory.

\section{Related Work}

\paragraph*{\textit{Robotic World Models}}
World models learn environment dynamics for planning and policy learning~\cite{ha2018world,hafner2019planet,hafner2025dreamerv3,hansen2023td}. Visual prediction represents anticipated interaction outcomes as videos: iVideoGPT models videos with discrete tokens~\cite{wu2024ivideogpt}, while RoboDreamer, IRASim, and Vid2World condition future generation on task descriptions or robot actions~\cite{zhou2024robodreamer,zhu2024irasim,huang2025vid2world}. These approaches draw on autoregressive and diffusion-based visual generation. VideoGPT models sequences of discrete video codes, VAR predicts successive spatial scales, and video diffusion models and diffusion transformers use iterative denoising~\cite{yan2021videogpt,tian2024var,ho2022videodiffusion,peebles2022dit}. Cosmos further develops world foundation models for physical AI~\cite{nvidia2025cosmos}. \roboimagine{} studies post-training GPT-style autoregressive robotic world models, whose discrete dynamics-token probabilities support candidate search and policy evaluation under shared observations, actions, and task instructions.

\paragraph*{\textit{Semantic and Visual Feedback}}
Reconstruction and perception guide visual prediction~\cite{mathieu2015deep,zhang2018unreasonable}, while VideoScore, VBench, WorldModelBench, and PhyCritic evaluate video semantics, dynamics, or physics~\cite{he2024videoscore,huang2024vbench,li2025worldmodelbench,xiong2026phycritic}. Trajectory preferences provide supervision~\cite{christiano2017preferences}; visual generation uses reward-weighted likelihood, policy gradients, and direct preference optimization~\cite{lee2023t2ifeedback,black2024ddpo,wallace2024diffusiondpo}. RLVR-World predicts future observations from video and actions using visual-metric rewards~\cite{wu2025rlvr}. RoboAlign-R1 distills a multimodal judge to score individual videos~\cite{wu2026roboalignr1}. Given shared observations, actions, and instructions, \roboimagine{} constructs futures by grouped diverse beam search. Pairwise task preferences from a frozen multimodal evaluator~\cite{somark2026mjev} and per-candidate pixel and perceptual scores jointly determine group-relative advantages for policy updates~\cite{shao2024deepseekmath}. We evaluate both candidate orders to mitigate position bias~\cite{zheng2023judging}.

\paragraph*{\textit{Candidate Search and Long-Horizon Memory}}
Diverse beam search discourages repeated decoding choices across groups~\cite{vijayakumar2016diverse}. Long-sequence generation additionally requires managing predicted histories. Diffusion Forcing combines causal modeling with token-wise noise levels, and Self Forcing trains on generated histories to address exposure bias~\cite{chen2024diffusionforcing,huang2025selfforcing}. StreamingLLM retains attention sinks and recent tokens; LongLive uses short-window attention and cache refresh, and LongLive-RAG retrieves relevant past latents~\cite{xiao2023efficient,yang2026longlive,hu2026longliverag}. RoboAlign-R1 periodically re-encodes visual context~\cite{wu2026roboalignr1}. \roboimagine{} retains a fixed initial context and dynamic anchor, together with branch-specific rolling state--action histories. It records and replays each candidate's history for consistent policy scoring, coupling bounded-context generation with diverse candidate comparison.

\section{Method}
\begin{figure*}[!t]
  \centering
  \includegraphics[width=\linewidth]{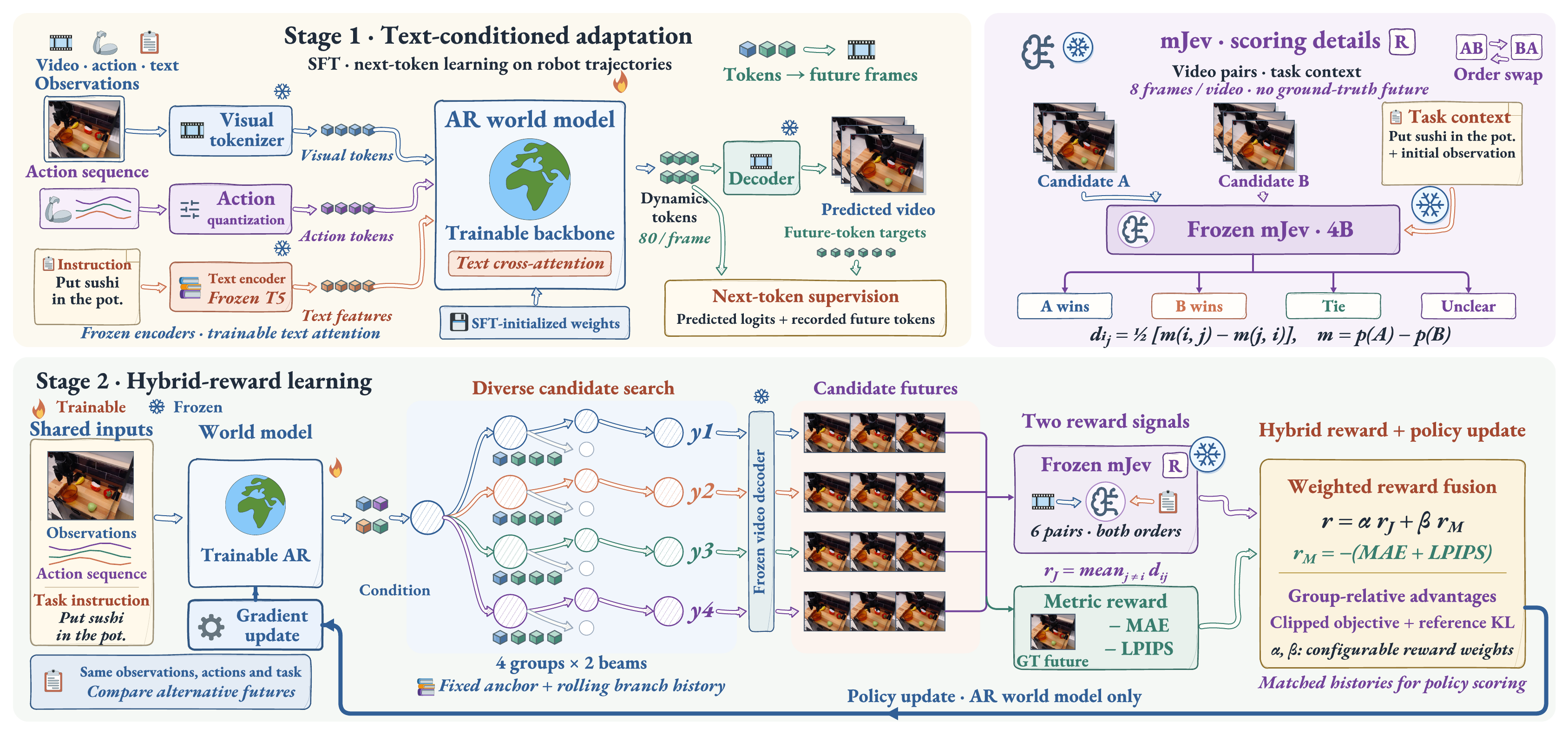}
  \caption{\textbf{Multimodal adaptation and learning from complementary feedback in \roboimagine{}.}
  Stage 1 learns dynamics-token prediction conditioned on observations, actions, and text. Stage 2 constructs diverse candidate futures with branch-specific rolling histories and combines semantic preferences from a frozen Jev evaluator with pixel- and perceptual-level visual feedback. Only the visual-feedback branch accesses recorded future frames during post-training. The generator and its text cross-attention receive policy gradients; the visual codec and feedback models remain frozen.}
  \label{fig:extension_hybrid_reward}
\end{figure*}

\begin{figure*}[!t]
  \centering
  \includegraphics[width=\linewidth]{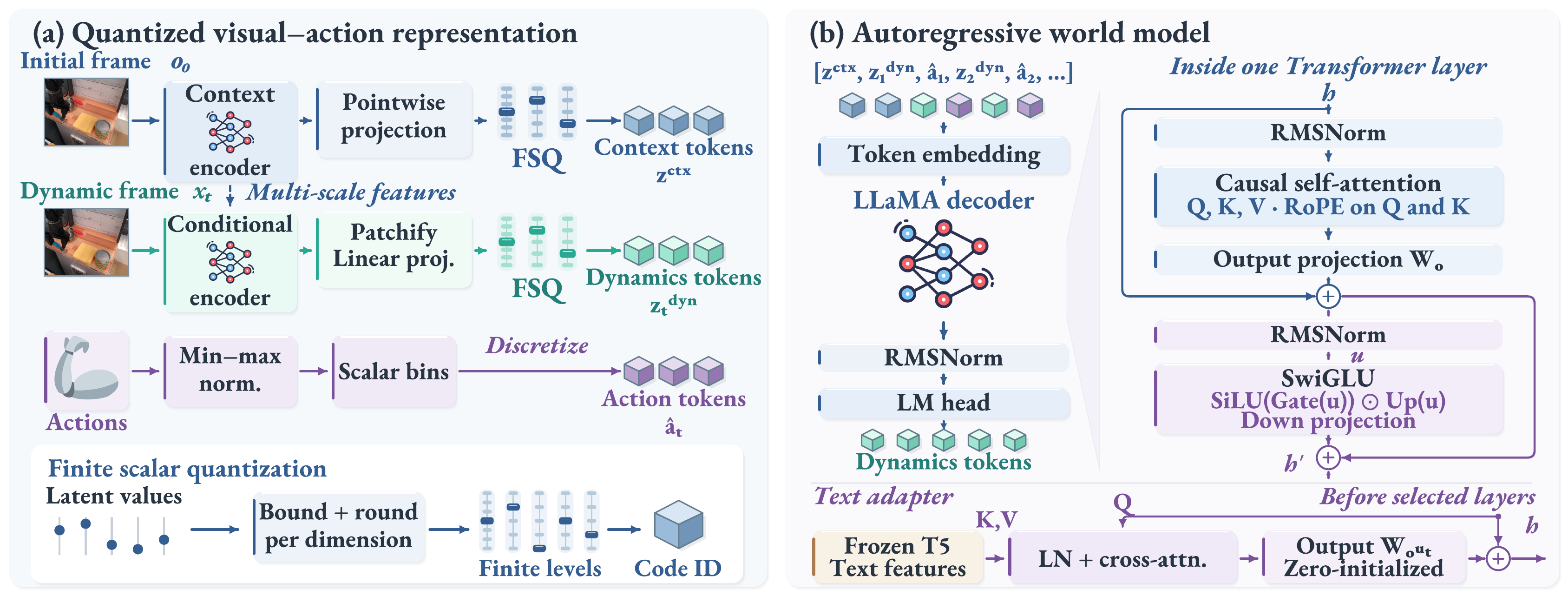}
  \caption{\textbf{Quantized representation and autoregressive prediction in \roboimagine{}.}
  \textbf{(a)}~Context-conditioned visual encoding produces 1,280 context tokens and 80 dynamics tokens per frame through finite scalar quantization (FSQ); discretized actions form the shared visual--action prefix. FSQ uses an implicit grid with 4,375 codes per visual branch.
  \textbf{(b)}~A 12-layer LLaMA decoder predicts dynamics tokens. The expanded layer shows causal attention, rotary positions, output projection, and the gated feed-forward block with residual connections. Text cross-attention is inserted before layers 4, 8, and 12; text features are not discrete prefix tokens.}
  \label{fig:model}
\end{figure*}

\begin{figure*}[!t]
  \centering
  \includegraphics[width=\textwidth]{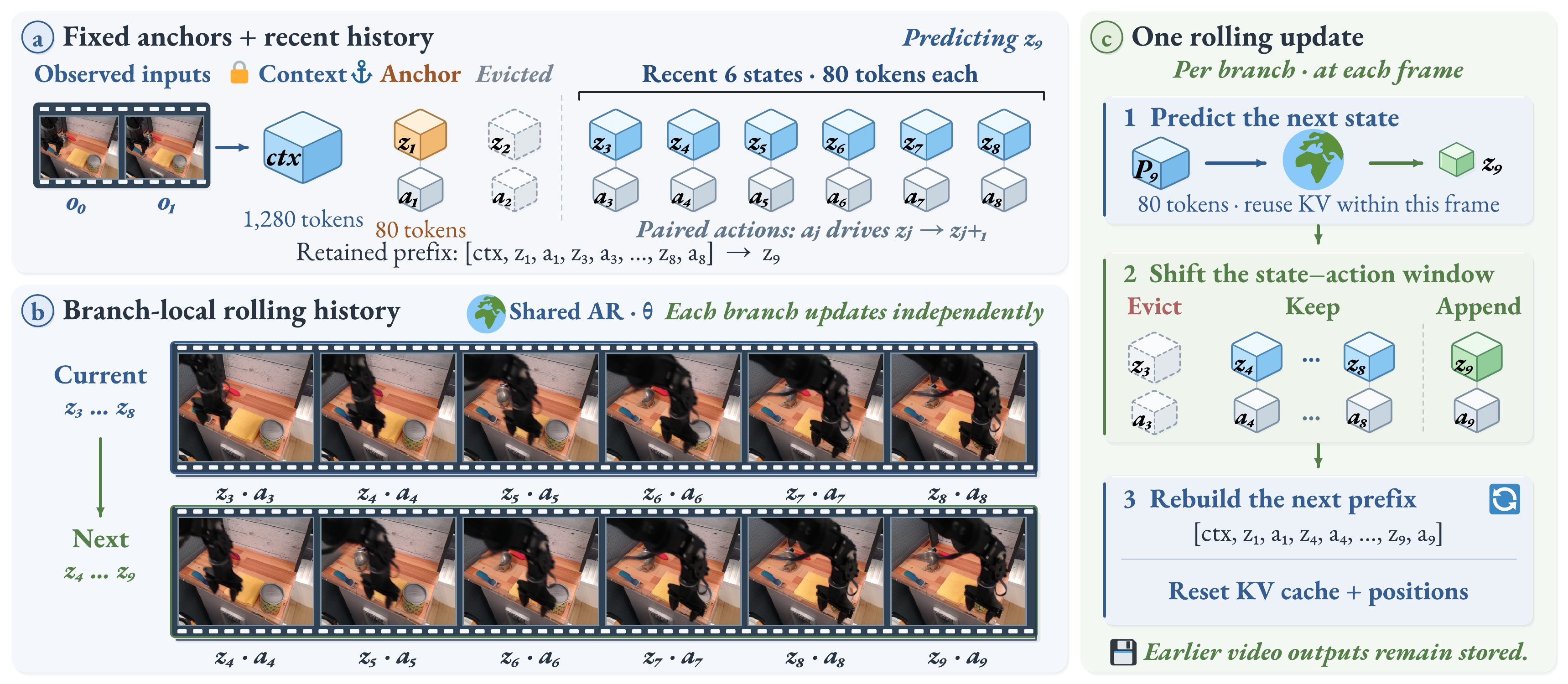}
  \caption{\textbf{Fixed-anchor branch memory for long-horizon generation in \roboimagine{}.} (a) The retained prefix combines fixed initial context and an observed dynamics anchor with six recent predicted states and paired actions. (b) A previously generated BridgeV2 rollout illustrates the shift from $z_{3:8}$ to $z_{4:9}$. Each search branch updates its own history using shared AR parameters. (c) The model predicts a new state, shifts the state--action window, and rebuilds its prefix and KV cache for the next frame. Both anchors and earlier video outputs remain stored. Cells represent token blocks; $z_j$ abbreviates $z_j^{\mathrm{dyn}}$ and $a_j$ denotes $\hat a_j$.}
  \label{fig:r2_branch_memory}
\end{figure*}

\begin{figure*}[!t]
  \centering
  \includegraphics[width=\textwidth]{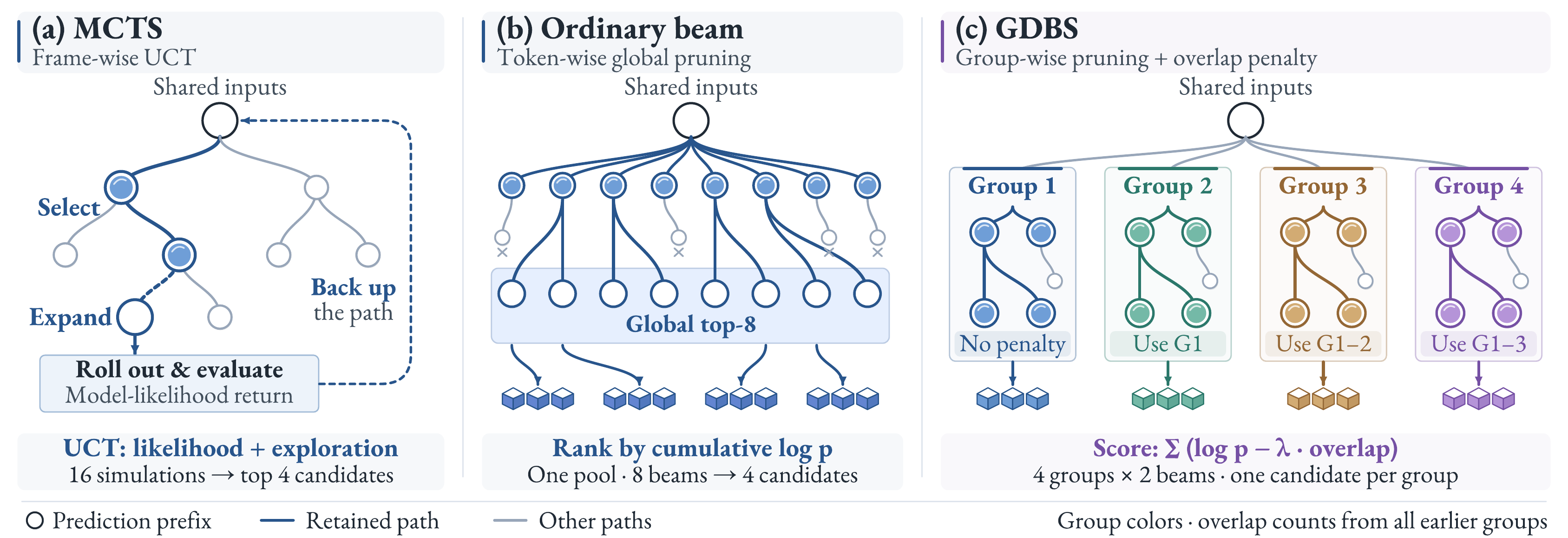}
  \caption{\textbf{Candidate construction with different search strategies.}
  (a) Likelihood-guided MCTS selects and expands frame prefixes, rolls out the remaining horizon, and backs up returns along the selected ancestor path.
  (b) Ordinary beam search ranks token prefixes in a single pool by cumulative model log probability.
  (c) GDBS maintains separate pools and penalizes tokens selected by all earlier groups at the same position, retaining one complete candidate per group. Colors identify groups; trees are schematic. The depicted search budgets differ.}
  \label{fig:search_comparison}
\end{figure*}

Figure~\ref{fig:extension_hybrid_reward} presents the overall framework. \roboimagine{} first establishes multimodal autoregressive prediction through supervised fine-tuning, then constructs candidate futures with GDBS and optimizes the model using complementary semantic and visual feedback. Each candidate maintains its own rolling history, which is replayed during policy evaluation to associate feedback with the actual prediction path.

\subsection{GPT-Style Autoregressive Robotic World Model}\label{sec:ar_world_model}

Given initial observations $o_0,o_1$, actions $a_{1:T}$, and an instruction $l$, the model predicts future visual states $x_{2:T+1}$. As shown in Figure~\ref{fig:model}, discrete visual--action tokens are processed by a causal Transformer with text cross-attention. Token probabilities provide a common interface for search and policy optimization.

\paragraph*{Visual--action representation}
A context-conditioned dual-branch encoder~\cite{wu2024ivideogpt} and finite scalar quantization (FSQ)~\cite{mentzer2024finite} encode the initial scene and subsequent states:
\begin{equation}\label{eq:visual_tokenization}
\begin{aligned}
z^{\mathrm{ctx}}&=Q_{\mathrm{ctx}}\!\left(E_{\mathrm{ctx}}(o_0)\right),\\
z_t^{\mathrm{dyn}}&=Q_{\mathrm{dyn}}\!\left(E_{\mathrm{dyn}}(x_t;o_0)\right).
\end{aligned}
\end{equation}
Here, $x_1=o_1$ and $t=1,\ldots,T+1$. The encoders include feature extraction and projection; the dynamics branch conditions on multiscale initial-scene features. Context and dynamics representations contain $N_c=1280$ and $N_d=80$ tokens, respectively.

Actions are normalized using training-set ranges and discretized dimension-wise into bins, yielding $\hat a_j=Q_a(a_j)$ for the transition from state $j$ to $j+1$. The resulting sequence is
\begin{equation}\label{eq:visual_action_sequence}
s=\bigl[z^{\mathrm{ctx}},z_1^{\mathrm{dyn}},\hat a_1,z_2^{\mathrm{dyn}},\hat a_2,\ldots\bigr].
\end{equation}
Predicting $z_t^{\mathrm{dyn}}$ therefore uses $a_{<t}$, with $a_{t-1}$ specifying the current transition. Modality-specific vocabulary offsets distinguish token types within the shared autoregressive backbone.

\paragraph*{Text conditioning and autoregressive prediction}
A frozen T5 encoder~\cite{raffel2020t5} produces instruction features $e_l=E_{\mathrm{text}}(l)$. Cross-attention at selected layers uses visual--action hidden states as queries and text features as keys and values. Its newly inserted output projections are zero-initialized. Let $P_t$ denote the retained visual--action prefix defined in Section~\ref{sec:search}. Frame tokens factorize as
\begin{equation}\label{eq:dynamics_factorization}
\pi_\theta\!\left(z_t^{\mathrm{dyn}}\mid P_t,e_l\right)
=\prod_{n=1}^{N_d}\pi_\theta\!\left(z_{t,n}^{\mathrm{dyn}}\mid P_t,z_{t,<n}^{\mathrm{dyn}},e_l\right).
\end{equation}
The distribution $\pi_\theta$ is renormalized over valid dynamics tokens. A frozen visual decoder reconstructs each completed frame from its dynamics tokens and the initial context.

\paragraph*{Supervised initialization}
Let $P_t^{\mathrm{GT}}$ contain recorded historical frames and actions, and let $z_{t,n}^{\mathrm{GT}}$ denote the target dynamics token. Supervised fine-tuning minimizes
\begin{equation}\label{eq:ar}
\mathcal L_{\mathrm{SFT}}=-\frac{1}{TN_d}\sum_{t=2}^{T+1}\sum_{n=1}^{N_d}
\log p_\theta\!\left(z_{t,n}^{\mathrm{GT}}\mid P_t^{\mathrm{GT}},z_{t,<n}^{\mathrm{GT}},e_l\right),
\end{equation}
where $p_\theta$ is the distribution over the full shared vocabulary, and loss applies only to target dynamics tokens. The visual codec and text encoder remain frozen; the backbone and text cross-attention are trained. The supervised model initializes post-training, and a frozen copy serves as the reference policy.

\subsection{Diverse Candidate Search}\label{sec:search}

Likelihood-based selection can concentrate candidates on similar prediction paths. GDBS balances model likelihood with cross-group differences, as illustrated alongside MCTS and ordinary beam search in Figure~\ref{fig:search_comparison}.

\paragraph*{Branch-specific history}
As shown in Figure~\ref{fig:r2_branch_memory}, the retained prefix $P_t$ contains the fixed initial context $z^{\mathrm{ctx}}$, the observed dynamics anchor $z_1^{\mathrm{dyn}}$, and each branch's latest $W$ predicted states with their paired actions, ending with the action for the next transition. Every generated frame is appended to the output, while the input history rolls forward. The window thus limits conditioning history, not output length. We use $W=6$, excluding the fixed anchor, and record the actual generation prefixes for policy evaluation.

\paragraph*{Grouped search}
Search uses $G$ groups with $B$ beams each. At dynamics-token position $u$, groups are processed sequentially. Let $\mathcal V_{h,u}$ be the multiset of final tokens from the $B$ extensions retained by earlier group $h$. The normalized overlap penalty is
\begin{equation}\label{eq:diversity_frequency}
f_{g,u}(v)=\begin{cases}
0,&g=1,\\[4pt]
\displaystyle\frac{\sum_{h<g}\sum_{v'\in\mathcal V_{h,u}}\mathbf1[v'=v]}{(g-1)B},&g>1.
\end{cases}
\end{equation}
Repeated tokens contribute repeated counts. Under the generating snapshot $\pi_{\theta_{\mathrm{old}}}$, extending prefix $y$ by token $v$ gives
\begin{equation}\label{eq:diverse_search}
\begin{aligned}
\mathcal S_{g,u}(yv)={}&\mathcal S_{g,u-1}(y)\\
&+\log\pi_{\theta_{\mathrm{old}}}(v\mid\mathcal H(y),e_l)
-\lambda_{\mathrm{div}}f_{g,u}(v).
\end{aligned}
\end{equation}
The history $\mathcal H(y)$ includes the retained visual--action prefix and preceding within-frame tokens. The penalty changes search scores, not policy normalization. Each group retains the top $B$ extensions across all parent beams and valid tokens, inheriting the selected parents' histories. We use $G=4$, $B=2$, and $\lambda_{\mathrm{div}}=0.1$, with fixed group membership and no length normalization.

\paragraph*{Complete candidates}
Each trajectory contains $L=TN_d$ future dynamics tokens. From the final beam set $\mathcal B_{g,L}$, we retain
\begin{equation}\label{eq:search_selection}
y^{(g)}=\underset{y\in\mathcal B_{g,L}}{\arg\max}\;\mathcal S_{g,L}(y),\qquad g=1,\ldots,G.
\end{equation}
The frozen codec decodes these trajectories:
\begin{equation}\label{eq:search_candidates}
\begin{aligned}
y^{(g)}&\equiv\hat z_{2:T+1}^{\mathrm{dyn},(g)},\\
\hat x_{2:T+1}^{(g)}&=\mathcal D_{\mathrm{vis}}(z^{\mathrm{ctx}},y^{(g)}).
\end{aligned}
\end{equation}
This produces one candidate per group, sharing external conditions while retaining branch-specific histories. We use $i$ to index the resulting candidates below.

\subsection{Complementary Semantic and Visual Feedback}\label{sec:reward}

Semantic feedback measures relative task progress, while visual feedback constrains prediction errors. Their fusion assigns a reward to each candidate trajectory.

\paragraph*{Semantic preference feedback}
A frozen multimodal evaluator $J_\phi$, implemented with the 4B mJev checkpoint (abbreviated as Jev)~\cite{somark2026mjev}, compares candidate pairs using instruction $l$, initial observation $o_1$, and eight temporally matched frames from each video. With frame-sampling operator $\mathcal F_8$, its output is
\begin{equation}\label{eq:jev_output}
\begin{aligned}
\mathbf q_{ij}&=J_\phi\!\left(l,o_1,\mathcal F_8(\hat{x}^{(i)}),\mathcal F_8(\hat{x}^{(j)})\right)\\
&=[q_{ij}^{\succ},q_{ij}^{\prec},q_{ij}^{=},q_{ij}^{?}].
\end{aligned}
\end{equation}
The four probabilities denote preference for the first or second candidate, comparable progress, and insufficient evidence. One forward pass reads four answer-token logits and normalizes them with softmax. The evaluator receives neither ground-truth future frames nor numerical actions.

To mitigate presentation-order bias, we evaluate both orders:
\begin{equation}\label{eq:bidirectional_preference}
m_{ij}=q_{ij}^{\succ}-q_{ij}^{\prec},\qquad d_{ij}=\tfrac12(m_{ij}-m_{ji}).
\end{equation}
Thus, $d_{ij}=-d_{ji}$, with positive values favoring candidate $i$. Tie and uncertainty probabilities remain in the normalization. The semantic reward is
\begin{equation}\label{eq:semantic_reward}
r_J^{(i)}=\frac{1}{G-1}\sum_{j\ne i}d_{ij}.
\end{equation}
Four candidates require six unordered pairs and twelve ordered comparisons.

\paragraph*{Visual consistency feedback}
For spatially and temporally aligned predictions and recorded futures, pixel and perceptual errors are
\begin{equation}\label{eq:visual_errors}
\begin{aligned}
e_{\mathrm{pix}}^{(i)}&=\frac{1}{T}\sum_{t=2}^{T+1}\frac{\|\hat{x}_t^{(i)}-x_t\|_1}{3H_{\mathrm{img}}W_{\mathrm{img}}},\\
e_{\mathrm{perc}}^{(i)}&=\frac{1}{T}\sum_{t=2}^{T+1}D_{\mathrm{LPIPS}}(\hat{x}_t^{(i)},x_t).
\end{aligned}
\end{equation}
RGB values lie in $[0,1]$; the frozen VGG-based LPIPS network~\cite{zhang2018unreasonable} receives inputs mapped to $[-1,1]$. We use equal error coefficients:
\begin{equation}\label{eq:visual_reward}
r_V^{(i)}=-\bigl(e_{\mathrm{pix}}^{(i)}+e_{\mathrm{perc}}^{(i)}\bigr).
\end{equation}
During post-training, ground-truth future frames enter only this feedback branch, not candidate-generation histories.

\paragraph*{Reward fusion}
To balance the feedback scales, each branch is standardized within the candidate set:
\begin{equation}\label{eq:reward_standardization}
\widetilde r_k^{(i)}=\frac{r_k^{(i)}-\mu_k}{\max(\sigma_k,\epsilon_{\mathrm{num}})},\qquad k\in\{J,V\},
\end{equation}
Here, $\mu_k$ and $\sigma_k$ are the candidate mean and sample standard deviation, with $\epsilon_{\mathrm{num}}=10^{-6}$. The fused reward is
\begin{equation}\label{eq:hybrid_reward}
r^{(i)}=w_J\widetilde r_J^{(i)}+w_V\widetilde r_V^{(i)}.
\end{equation}
We use $(w_J,w_V)=(0.25,0.75)$ for combined feedback. All rewards and normalization statistics are detached from the generator's gradient graph.

\subsection{Search-Guided Policy Optimization}\label{sec:policy_optimization}

Group-relative advantages~\cite{shao2024deepseekmath} adjust candidate probabilities under their recorded generation histories, with regularization toward the frozen supervised policy.

\paragraph*{Group-relative advantages}
Writing $r_i=r^{(i)}$, let $\mu_r$ and $\sigma_r$ be the mean and sample standard deviation of fused rewards. We define
\begin{equation}\label{eq:r2_advantage}
\widehat A_i=\begin{cases}
\displaystyle\frac{r_i-\mu_r}{\sigma_r+\epsilon_{\mathrm{num}}},&\sigma_r>\epsilon_{\mathrm{num}},\\[4pt]
0,&\text{otherwise}.
\end{cases}
\end{equation}
All tokens in a candidate share its detached advantage. Branch standardization balances the feedback sources, whereas advantage normalization scales the resulting fused signal.

\paragraph*{History-consistent evaluation}
For token $y_{i,u}$, let $\mathcal H_{i,u}$ contain its actual retained prefix and preceding within-frame tokens. The policy ratio is
\begin{equation}\label{eq:r2_ratio}
\rho_{i,u}(\theta)=\frac{\pi_\theta(y_{i,u}\mid\mathcal H_{i,u},e_l)}{\pi_{\theta_{\mathrm{old}}}(y_{i,u}\mid\mathcal H_{i,u},e_l)}.
\end{equation}
Candidates and histories remain fixed during an update, and the denominator is detached. Both policies normalize over valid dynamics tokens without the search penalty.

\paragraph*{Regularized policy objective}
The clipped surrogate~\cite{schulman2017ppo} is
\begin{equation}\label{eq:r2_clipped_surrogate}
\begin{aligned}
s_{i,u}=\min\bigl\{&\rho_{i,u}\widehat A_i,\\
&\operatorname{clip}(\rho_{i,u},1-\eta,1+\eta)\widehat A_i\bigr\}.
\end{aligned}
\end{equation}
For $\widehat A_i<0$, dual clipping gives $\widetilde s_{i,u}=\max(s_{i,u},c_{\mathrm{dual}}\widehat A_i)$; otherwise $\widetilde s_{i,u}=s_{i,u}$. We set $\eta=0.2$ and $c_{\mathrm{dual}}=3$. With the frozen supervised policy $\pi_{\mathrm{ref}}$, reference regularization uses
\begin{equation}\label{eq:r2_reference_regularizer}
\begin{aligned}
\Delta_{i,u}&=\log\frac{\pi_{\mathrm{ref}}(y_{i,u}\mid\mathcal H_{i,u},e_l)}{\pi_\theta(y_{i,u}\mid\mathcal H_{i,u},e_l)},\\
k_{i,u}&=\min\{\exp(\Delta_{i,u})-\Delta_{i,u}-1,\kappa\}.
\end{aligned}
\end{equation}
The objective is
\begin{equation}\label{eq:r2_policy_objective}
\mathcal L_{\mathrm{policy}}=\frac1{GL}\sum_{i=1}^G\sum_{u=1}^L
\bigl[-\widetilde s_{i,u}+\lambda_{\mathrm{KL}}k_{i,u}\bigr],
\end{equation}
where $\lambda_{\mathrm{KL}}=10^{-3}$ and $\kappa=10$. This capped, selected-token penalty is KL-style regularization, not an unbiased full-distribution KL estimate. Only the backbone and text cross-attention are updated.

\paragraph*{Actual update and evaluation}
Each candidate batch receives one optimizer step, with gradients accumulated across frame blocks. At $\theta=\theta_{\mathrm{old}}$, ratios equal one and clipping is inactive. The reward component then has the same gradient as
\begin{equation}\label{eq:r2_weighted_likelihood}
\mathcal L_{\mathrm{weighted}}=-\frac1{GL}\sum_{i=1}^G\sum_{u=1}^L
\widehat A_i\log\pi_\theta(y_{i,u}\mid\mathcal H_{i,u},e_l).
\end{equation}
Positive and negative advantages respectively favor increasing and decreasing candidate-token probabilities; shared parameters and reference regularization jointly determine the update. Candidates are regenerated after each step, and an all-tied global batch is skipped. This is an empirical surrogate over search-selected candidates without importance correction for their selection. Training uses GDBS; evaluation uses frame-wise ordinary beam search of width four, retaining one completed frame before continuing, without an evaluator or ground-truth futures.

\subsection{Candidate Differences and Comparative Learning}\label{sec:r2_analysis}

Consider $G\geq2$ fixed candidates at $\theta_{\mathrm b}=\theta_{\mathrm{old}}$, holding their tokens, histories, text features, rewards, and statistics constant. Define
\begin{equation}\label{eq:r2_candidate_score}
\begin{aligned}
\ell_i(\theta)&=\frac1L\sum_{u=1}^{L}\log\pi_\theta(y_{i,u}\mid\mathcal H_{i,u},e_l),\\
\mathbf v_i&=\left.\nabla_\theta\ell_i(\theta)\right|_{\theta=\theta_{\mathrm b}}.
\end{aligned}
\end{equation}
Let $s_r=\sigma_r+\epsilon_{\mathrm{num}}$, and assume selected-token probabilities are positive and differentiable.

\begin{lemma}\label{lem:r2_pairwise_gradient}
At $\theta_{\mathrm b}$, where clipping is inactive, the reward-ascent gradient, excluding reference regularization, satisfies
\begin{equation}\label{eq:r2_pairwise_gradient}
\begin{aligned}
\mathbf g&=\frac1{Gs_r}\sum_{i=1}^G(r_i-\mu_r)\mathbf v_i\\
&=\frac1{G^2s_r}\sum_{i<j}(r_i-r_j)(\mathbf v_i-\mathbf v_j),
\end{aligned}
\end{equation}
provided $\sigma_r>\epsilon_{\mathrm{num}}$; otherwise $\mathbf g=\mathbf0$.
\end{lemma}

\begin{proof}[Proof sketch]
At the snapshot, $\rho_{i,u}=1$ with a detached denominator, so Eq.~\ref{eq:r2_weighted_likelihood} gives $\mathbf g=G^{-1}\sum_i\widehat A_i\mathbf v_i$. Expanding the pairwise sum yields
\[
\sum_{i<j}(r_i-r_j)(\mathbf v_i-\mathbf v_j)
=G\sum_i(r_i-\mu_r)\mathbf v_i.
\]
The complete proof appears in Appendix~\ref{app:r2_comparison_analysis}.
\end{proof}

\begin{corollary}\label{cor:r2_degenerate_feedback}
The reward-driven gradient vanishes if all candidate rewards coincide or all $\mathbf v_i$ coincide. Identical candidates with identical histories are a special case; reference regularization may still produce an update.
\end{corollary}

Equation~\ref{eq:r2_pairwise_gradient} shows that feedback differences weight differences in candidate score gradients: search constructs distinct paths, and feedback determines their relative update weights. The identity requires no independent-sampling assumption. Token differences alone do not guarantee a nonzero gradient, since pairwise contributions can cancel. Controlled ablations assess the empirical benefit; sufficient nondegeneracy conditions and a local fixed-candidate analysis are provided in Appendix~\ref{app:r2_comparison_analysis}.

\setlength{\floatsep}{10pt plus 2pt minus 2pt}
\setlength{\textfloatsep}{12pt plus 2pt minus 2pt}
\setlength{\dblfloatsep}{12pt plus 2pt minus 2pt}
\setlength{\dbltextfloatsep}{20pt plus 2pt minus 2pt}
\setlength{\intextsep}{10pt plus 2pt minus 2pt}
\section{Experiments}

We compare \roboimagine{} with existing methods on visual prediction quality and task-level consistency, then assess comparative post-training against continued supervised fine-tuning from the same initialization. Ablation studies examine feedback composition and candidate-search strategies. We further analyze long-horizon prediction and the effect of history-window size on quality and inference cost.

\subsection{Experimental Setup}\label{sec:experimental_setup}

\textbf{Datasets and prediction protocol.} We evaluate on RT-1~\cite{brohan2023rt1}, BridgeV2~\cite{walke2023bridgedata}, and RoboCasa~\cite{nasiriany2024robocasa}, covering real-world robot manipulation and simulated household tasks. Models are trained separately for each dataset. The main prediction evaluation uses 128 trajectories per dataset, except for the explicitly labeled 32-trajectory BridgeV2 validation result in Table~\ref{tab:robotworldbench_lowlevel}, and generates 32 future frames from initial observations and action sequences at $256\times320$ resolution. Generated predictions update the rollout history; future ground-truth images are never used for generation.

\textbf{Baselines.} The models in Tables~\ref{tab:robotworldbench_main} and~\ref{tab:robotworldbench_lowlevel} span five categories:

\noindent\textcolor{mydarkblue}{\ding{172}}~\textbf{Closed-source video generators:} Kling~2.6~\cite{kling2025video26}, Runway Gen-4.5~\cite{runway2025gen45}, MiniMax Hailuo~02~\cite{minimax2025hailuo02}, and Luma Dream Machine~\cite{luma2024dreammachine}.

\noindent\textcolor{codegreen!60!black}{\ding{173}}~\textbf{Open-source video generators:} HunyuanVideo-I2V~\cite{tencent2025hunyuanvideoi2v}, LTX-Video~\cite{hacohen2024ltxvideo}, Stable Video Diffusion XT~\cite{blattmann2023stablevideo}, Mochi-1~\cite{genmo2024mochi}, I2VGen-XL~\cite{zhang2023i2vgenxl}, CogVideoX-I2V~\cite{yang2024cogvideox}, OpenSora-I2V~\cite{zheng2024opensora}, and OpenSora-Plan-I2V~\cite{lin2024opensora_plan}.

\noindent\textcolor{codepurple}{\ding{174}}~\textbf{Action-conditioned and interactive world models:} iVideoGPT~\cite{wu2024ivideogpt}, RoboDreamer~\cite{zhou2024robodreamer}, IRASim~\cite{zhu2024irasim}, Vid2World~\cite{huang2025vid2world}, and DriftWorld~\cite{lu2026driftworld}.

\noindent\textcolor{red!70!black}{\ding{175}}~\textbf{RL-post-trained world models:} RLVR-World~\cite{wu2025rlvr} and RoboAlign-R1~\cite{wu2026roboalignr1}, which provide comparisons with alternative reward-alignment methods.

\noindent\textcolor{mydarkblue}{\ding{176}}~\textbf{Task-adapted video foundation models:} Wan2.2-TI2V-5B~\cite{wan2025wan22}, including the LoRA-adapted version in Table~\ref{tab:robotworldbench_main}; Cosmos-Predict2.5-2B~\cite{nvidia2025cosmos25}; and DreamDojo-2B~\cite{gao2026dreamdojo}.

Figure~\ref{fig:baseline_qualitative} additionally includes Ctrl-World~\cite{guo2026ctrlworld} and WEAVER~\cite{jain2026weaver} as qualitative baselines. External baselines use previously reported or adaptation-based evaluation results, with potentially different training budgets, conditioning interfaces, and decoding procedures. Controlled experiments share the initialization, test trajectories, and evaluation decoder to compare continued supervised fine-tuning, feedback combinations, and candidate-search strategies.

\textbf{Implementation and post-training.} \roboimagine{} starts from dataset-specific SFT weights. The controlled RT-1 experiments share a 50,000-step SFT checkpoint and perform 500 additional updates using AdamW with a learning rate of $10^{-6}$ and a global batch size of eight. Each condition generates ten-frame candidate futures. By default, GDBS uses four groups with two beams each and retains one complete candidate per group; the diversity penalty is $0.1$. Semantic and visual feedback have weights $0.25$ and $0.75$, respectively. Only the autoregressive generator and text cross-attention modules are updated; the visual codec, text encoder, and semantic evaluator remain frozen. Default memory retains the fixed anchors and six recent states. Evaluation uses frame-wise beam search with width four.

\textbf{Evaluation metrics.} We report PSNR, SSIM~\cite{wang2004image}, and LPIPS~\cite{zhang2018unreasonable} for visual prediction quality, and additionally use MAE in the training-dynamics analysis. Main prediction metrics are averaged over the 32 predicted frames within each trajectory and then across test trajectories. Task semantics and physical consistency are assessed with the six-dimensional RoboAlign-Judge rubric~\cite{wu2026roboalignr1}: instruction following, manipulation success, action--outcome consistency, temporal consistency, contact realism, and physics adherence, summing to a maximum of ten points. Additional multimodal evaluators score paired SFT and post-trained predictions; sample coverage is specified in the corresponding experiments. The independent multimodal evaluations use models distinct from the training reward model; the search comparison additionally reports feedback from the frozen training evaluator. We measure computational efficiency through training GPU-hours, generation latency, and peak GPU memory usage.

\subsection{Comparison with State-of-the-Art Methods}\label{sec:rq1}

\begin{table*}[!tp]
  \centering
  \normalsize
  \renewcommand{\scorestd}[2]{#1$_{\pm#2}$}
  \renewcommand{\arraystretch}{1.05}
  \setlength{\tabcolsep}{5.5pt}
  \setlength{\abovecaptionskip}{5pt}

  \caption{Robot world-model performance on RT-1 from \textsc{RobotWorldBench}, evaluated with RoboAlign-Judge (mean $\pm$ std.). Superscripts 1, 2, and 3 indicate the top three distinct model scores in each column; ties share a rank.}
  \label{tab:robotworldbench_main}
  \begin{fullwidthcomparison}{8}{lccccccc}
    \toprule
    \rowcolor{tblheadbg}
    Method & \multicolumn{3}{c}{Task Alignment} & \multicolumn{3}{c}{Physical Realism} & Total \\
    \cmidrule(lr){2-4} \cmidrule(lr){5-7}
    \rowcolor{tblheadbg}
     & Instr. $\uparrow$ & Manip. $\uparrow$ & Act.-Out. $\uparrow$ & Temp. $\uparrow$ & Contact $\uparrow$ & Phys. $\uparrow$ & $\uparrow$ \\
    \midrule
    \rowcolor{tblstripe}
    \textit{Real videos}             & 3.0 & 2.0 & 1.0 & 1.0 & 1.0 & 2.0 & 10.0 \\
    \midrule
    \multicolumn{8}{l}{\textcolor{red!70!black}{\textbf{\textit{Closed video models}}}} \\
    \rowcolor{tblstripe}
    Kling 2.6             & \plainscore{2.42}{0.09} & \plainscore{1.38}{0.10} & \plainscore{0.46}{0.08} & \plainscore{0.58}{0.07} & \plainscore{0.82}{0.06} & \plainscore{1.18}{0.11} & \plainscore{6.84}{0.22} \\
    Runway Gen-4.5        & \plainscore{2.34}{0.10} & \plainscore{1.29}{0.09} & \plainscore{0.43}{0.07} & \plainscore{0.55}{0.08} & \plainscore{0.79}{0.06} & \plainscore{1.12}{0.10} & \plainscore{6.52}{0.24} \\
    \rowcolor{tblstripe}
    MiniMax Hailuo 02     & \plainscore{2.18}{0.11} & \plainscore{1.17}{0.10} & \plainscore{0.38}{0.07} & \plainscore{0.47}{0.09} & \plainscore{0.71}{0.08} & \plainscore{1.01}{0.12} & \plainscore{5.92}{0.28} \\
    Luma Dream Machine    & \plainscore{2.03}{0.12} & \plainscore{1.08}{0.11} & \plainscore{0.35}{0.08} & \plainscore{0.44}{0.09} & \plainscore{0.69}{0.08} & \plainscore{0.92}{0.12} & \plainscore{5.51}{0.30} \\
    \midrule
    \multicolumn{8}{l}{\textcolor{codegreen!60!black}{\textbf{\textit{Open video models}}}} \\
    \rowcolor{tblstripe}
    HunyuanVideo-I2V      & \plainscore{1.20}{0.28} & \plainscore{0.40}{0.22} & \plainscore{0.24}{0.08} & \plainscore{0.56}{0.15} & \thirdscore{0.96}{0.08} & \thirdscore{1.56}{0.15} & \plainscore{4.92}{0.63} \\
    LTX-Video             & \plainscore{2.28}{0.07} & \plainscore{1.32}{0.07} & \plainscore{0.52}{0.07} & \plainscore{0.26}{0.05} & \plainscore{0.28}{0.10} & \plainscore{1.00}{0.18} & \plainscore{5.66}{0.36} \\
    \rowcolor{tblstripe}
    Stable Video Diffusion XT & \plainscore{1.84}{0.15} & \plainscore{0.98}{0.13} & \plainscore{0.46}{0.08} & \plainscore{0.06}{0.08} & \plainscore{0.36}{0.20} & \plainscore{0.58}{0.20} & \plainscore{4.28}{0.74} \\
    Mochi-1               & \plainscore{1.83}{0.12} & \plainscore{0.91}{0.10} & \plainscore{0.31}{0.07} & \plainscore{0.39}{0.08} & \plainscore{0.61}{0.07} & \plainscore{0.82}{0.11} & \plainscore{4.87}{0.29} \\
    \rowcolor{tblstripe}
    I2VGen-XL             & \plainscore{1.56}{0.15} & \plainscore{0.66}{0.14} & \plainscore{0.32}{0.04} & \plainscore{0.00}{0.00} & \plainscore{0.44}{0.19} & \plainscore{0.56}{0.16} & \plainscore{3.54}{0.57} \\
    CogVideoX-I2V         & \plainscore{1.94}{0.21} & \plainscore{1.00}{0.14} & \plainscore{0.40}{0.18} & \plainscore{0.42}{0.16} & \plainscore{0.92}{0.07} & \plainscore{1.24}{0.19} & \plainscore{5.92}{0.75} \\
    \rowcolor{tblstripe}
    OpenSora-I2V          & \plainscore{1.58}{0.13} & \plainscore{0.82}{0.11} & \plainscore{0.27}{0.06} & \plainscore{0.31}{0.07} & \plainscore{0.53}{0.08} & \plainscore{0.71}{0.10} & \plainscore{4.22}{0.30} \\
    OpenSora-Plan-I2V     & \plainscore{1.79}{0.12} & \plainscore{0.89}{0.10} & \plainscore{0.30}{0.07} & \plainscore{0.37}{0.08} & \plainscore{0.60}{0.08} & \plainscore{0.81}{0.11} & \plainscore{4.76}{0.29} \\
    \midrule
    \multicolumn{8}{l}{\textcolor{codepurple}{\textbf{\textit{Embodied / interactive world-model baselines}}}} \\
    \rowcolor{tblstripe}
    RLVR-World            & \plainscore{2.29}{0.10} & \plainscore{1.31}{0.09} & \plainscore{0.44}{0.07} & \plainscore{0.43}{0.08} & \plainscore{0.84}{0.06} & \plainscore{1.19}{0.10} & \plainscore{6.54}{0.23} \\
    iVideoGPT             & \thirdscore{2.60}{0.11} & \thirdscore{1.60}{0.11} & \thirdscore{0.70}{0.11} & \thirdscore{0.74}{0.14} & \plainscore{0.56}{0.10} & \plainscore{1.54}{0.15} & \thirdscore{7.74}{0.62} \\
    \rowcolor{tblstripe}
    RoboDreamer           & \plainscore{2.02}{0.16} & \plainscore{1.02}{0.16} & \plainscore{0.34}{0.05} & \plainscore{0.32}{0.12} & \plainscore{0.04}{0.05} & \plainscore{0.70}{0.17} & \plainscore{4.44}{0.56} \\
    Vid2World             & \plainscore{2.18}{0.10} & \plainscore{1.08}{0.04} & \plainscore{0.40}{0.09} & \plainscore{0.22}{0.07} & \secondscore{0.98}{0.04} & \plainscore{1.04}{0.14} & \plainscore{5.90}{0.30} \\
    \rowcolor{tblstripe}
    RoboAlign-R1 & \secondscore{2.72}{0.06} & \secondscore{1.72}{0.07} & \secondscore{0.72}{0.05} & \secondscore{0.78}{0.05} & \bestscore{1.00}{0.04} & \secondscore{1.58}{0.07} & \secondscore{8.52}{0.15} \\
    \midrule
    \multicolumn{8}{l}{\textcolor{codepurple}{\textbf{\textit{Additional training baselines}}}} \\
    \rowcolor{tblstripe}
    Wan2.2-TI2V-5B (LoRA) & \plainscore{2.40}{0.09} & \plainscore{1.02}{0.11} & \plainscore{0.41}{0.08} & \plainscore{0.24}{0.10} & \secondscore{0.98}{0.05} & \plainscore{0.96}{0.12} & \plainscore{6.01}{0.24} \\
    \midrule
    \rowcolor{mydarkblue!6}
    \textcolor{mydarkblue}{\textbf{\roboimagine{} (ours)}} & \bestscore{2.82}{0.02} & \bestscore{1.78}{0.05} & \bestscore{0.82}{0.03} & \bestscore{0.79}{0.03} & \bestscore{1.00}{0.01} & \bestscore{1.77}{0.04} & \bestscore{8.98}{0.10} \\
    \bottomrule
  \end{fullwidthcomparison}

\end{table*}

\afterpage{%

\begin{table*}[!t]
  \centering
  \normalsize
  \renewcommand{\arraystretch}{1.05}
  \setlength{\tabcolsep}{5.5pt}
  \caption{Video prediction over 32 frames. Superscripts rank the reported values, as in Table~\ref{tab:robotworldbench_main}. \roboimagine{}'s BridgeV2 entry reports the best checkpoint selected by LPIPS on 32 validation trajectories; other entries use 128 trajectories. Rankings do not imply matched-cohort comparisons.}
  \label{tab:robotworldbench_lowlevel}
  \begin{fullwidthcomparison}{10}{lccccccccc}
    \toprule
    \rowcolor{tblheadbg}
    Method & \multicolumn{3}{c}{RT-1} & \multicolumn{3}{c}{BridgeV2} & \multicolumn{3}{c}{RoboCasa} \\
    \cmidrule(lr){2-4} \cmidrule(lr){5-7} \cmidrule(lr){8-10}
    \rowcolor{tblheadbg}
     & PSNR $\uparrow$ & SSIM $\uparrow$ & LPIPS $\downarrow$ & PSNR $\uparrow$ & SSIM $\uparrow$ & LPIPS $\downarrow$ & PSNR $\uparrow$ & SSIM $\uparrow$ & LPIPS $\downarrow$ \\
    \midrule
    \multicolumn{10}{l}{\textcolor{codegreen!60!black}{\textbf{\textit{Video Prediction Baselines}}}} \\
    \rowcolor{tblstripe}
    iVideoGPT & 15.96 & 0.5571 & 0.3833 & \thirdmean{20.06} & \thirdmean{0.7791} & 0.1899 & \thirdmean{16.67} & \thirdmean{0.7187} & 0.2366 \\
    \midrule
    \multicolumn{10}{l}{\textcolor{codepurple}{\textbf{\textit{Action-Conditioned World Models}}}} \\
    \rowcolor{tblstripe}
    IRASim & 16.83 & 0.5936 & 0.3474 & 17.39 & 0.7094 & 0.2524 & 13.55 & 0.5834 & 0.3706 \\
    Vid2World & 19.03 & 0.7210 & 0.1960 & 18.92 & 0.7110 & 0.1880 & 15.40 & 0.7000 & \thirdmean{0.2287} \\
    \rowcolor{tblstripe}
    DriftWorld & \secondmean{20.54} & \thirdmean{0.7410} & 0.2595 & \secondmean{20.32} & 0.7429 & 0.2379 & 16.50 & \secondmean{0.7200} & 0.2891 \\
    \midrule
    \multicolumn{10}{l}{\textcolor{red!70!black}{\textbf{\textit{RL-Based World Models}}}} \\
    \rowcolor{tblstripe}
    RLVR-World & 19.47 & 0.7360 & \thirdmean{0.1820} & 19.06 & 0.7210 & \thirdmean{0.1780} & 15.80 & 0.7100 & \secondmean{0.2017} \\
    RoboAlign-R1 & \thirdmean{19.72} & \secondmean{0.7450} & \secondmean{0.1730} & 20.00 & 0.7310 & \secondmean{0.1750} & \secondmean{16.71} & 0.7019 & 0.2319 \\
    \midrule
    \multicolumn{10}{l}{\textcolor{codepurple}{\textbf{\textit{Adapted Video Foundation Models}}}} \\
    \rowcolor{tblstripe}
    Wan2.2-TI2V-5B & 18.42 & 0.7058 & 0.2478 & 16.35 & 0.6889 & 0.2568 & 14.37 & 0.6906 & 0.2711 \\
    Cosmos-Predict2.5-2B & 17.77 & 0.6765 & 0.2889 & 18.98 & \secondmean{0.7799} & 0.1870 & 14.30 & 0.6802 & 0.2916 \\
    \rowcolor{tblstripe}
    DreamDojo-2B & 18.47 & 0.7152 & 0.2660 & 16.95 & 0.7461 & 0.2563 & 13.92 & 0.6349 & 0.3278 \\
    \midrule
    \rowcolor{mydarkblue!6}
    \textcolor{mydarkblue}{\textbf{\roboimagine{} (ours)}} & \bestmean{22.57} & \bestmean{0.8035} & \bestmean{0.1541} & \bestmean{21.53} & \bestmean{0.7880} & \bestmean{0.1527} & \bestmean{18.39} & \bestmean{0.7727} & \bestmean{0.1768} \\
    \bottomrule
  \end{fullwidthcomparison}

\end{table*}
%
\begin{figure*}[!t]
  \centering
  \includegraphics[width=\textwidth]{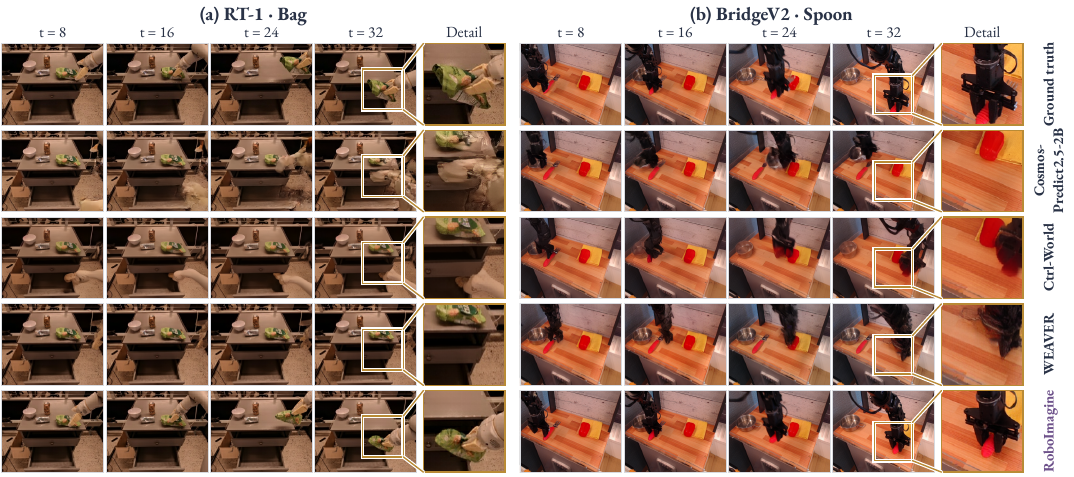}
  \caption{\textbf{Qualitative comparison on RT-1 and BridgeV2.} Selected bag- and spoon-manipulation examples at steps 8, 16, 24, and 32. Gold boxes mark the shared step-32 regions enlarged on the right.}
  \label{fig:baseline_qualitative}
\end{figure*}
%
\begin{figure*}[!t]
  \centering
  \includegraphics[width=\textwidth]{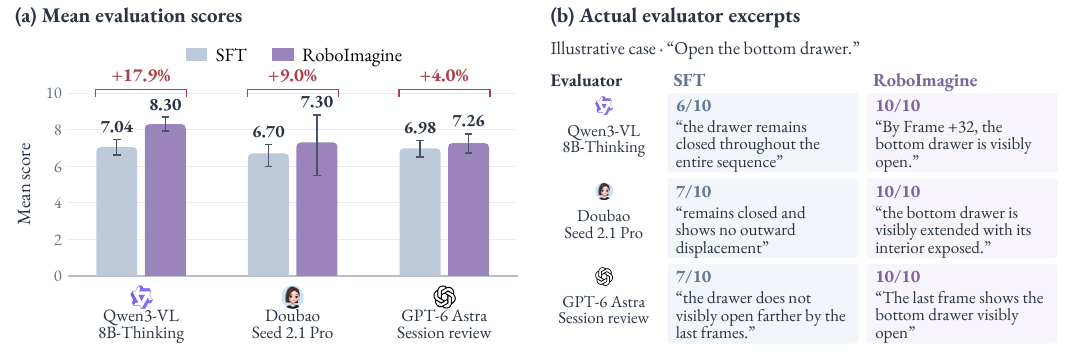}
  \caption{\textbf{Multimodal evaluation and illustrative feedback.} (a) Mean scores with 95\% bootstrap confidence intervals on 97, 10, and 50 paired trajectories, respectively. (b) Verbatim evaluator excerpts and case-specific scores for one RT-1 trajectory (case 024). GPT-6 Astra denotes a configured-model session review, not a version-pinned API evaluation.}
  \label{fig:r2_multijudge_mean}
\end{figure*}
}

\textbf{Task alignment and physical realism.} Table~\ref{tab:robotworldbench_main} evaluates task-level behavior beyond visual similarity. \roboimagine{} achieves the highest total score among the compared models, reaching $8.98\pm0.10$ out of ten, compared with $8.52\pm0.15$ for RoboAlign-R1. It ranks first in five dimensions and ties for first in contact realism. In particular, action--outcome consistency increases from $0.72$ to $0.82$, and physics adherence from $1.58$ to $1.77$. These results indicate that the gains extend beyond appearance to the consistency of action-conditioned interactions and outcomes.

\textbf{Visual prediction quality.} Table~\ref{tab:robotworldbench_lowlevel} reports visual prediction metrics across the three datasets. On RT-1 and RoboCasa, \roboimagine{} achieves PSNR values of $22.57$ and $18.39$~dB, exceeding the strongest listed baseline by $2.03$ and $1.68$~dB, respectively. LPIPS decreases by $10.9\%$ and $12.3\%$ relative to the lowest listed baseline value, alongside higher SSIM. On the 32 BridgeV2 validation trajectories, the best checkpoint selected by validation LPIPS reaches $21.53$~dB PSNR, $0.7880$ SSIM, and $0.1527$ LPIPS, compared with $21.39$~dB, $0.7865$, and $0.1539$ for its SFT initialization. Its reported values are the best listed in the BridgeV2 columns, although the differing cohorts preclude a matched-cohort comparison. On the separate 128-trajectory reporting cohort, the same checkpoint gives $21.65$~dB PSNR, $0.7926$ SSIM, and $0.1467$ LPIPS versus $21.74$~dB, $0.7930$, and $0.1460$ for SFT, so its validation gain does not establish a gain on that cohort. External baselines also differ in training and conditioning settings; the shared-initialization comparison below separately examines the effect of post-training.

\textbf{Qualitative comparison.} Figure~\ref{fig:baseline_qualitative} shows selected, matched trajectories at four prediction steps. In the RT-1 example, \roboimagine{} follows the gripper--bag interaction more closely, whereas several baselines leave the bag on the counter or distort the robot and object. In the BridgeV2 spoon-manipulation example, it better preserves the gripper geometry and its spatial relationship to the red spoon. Cosmos leaves the spoon on the work surface, while Ctrl-World and WEAVER show larger deviations in gripper position or shape. The shared step-32 crops make these local differences more visible. The RT-1 example uses the 500-update combined-feedback model; the BridgeV2 example uses the same validation-selected checkpoint as Table~\ref{tab:robotworldbench_lowlevel}. These qualitative examples complement the aggregate evaluation.

\begin{figure*}[!t]
  \centering
  \includegraphics[width=\textwidth]{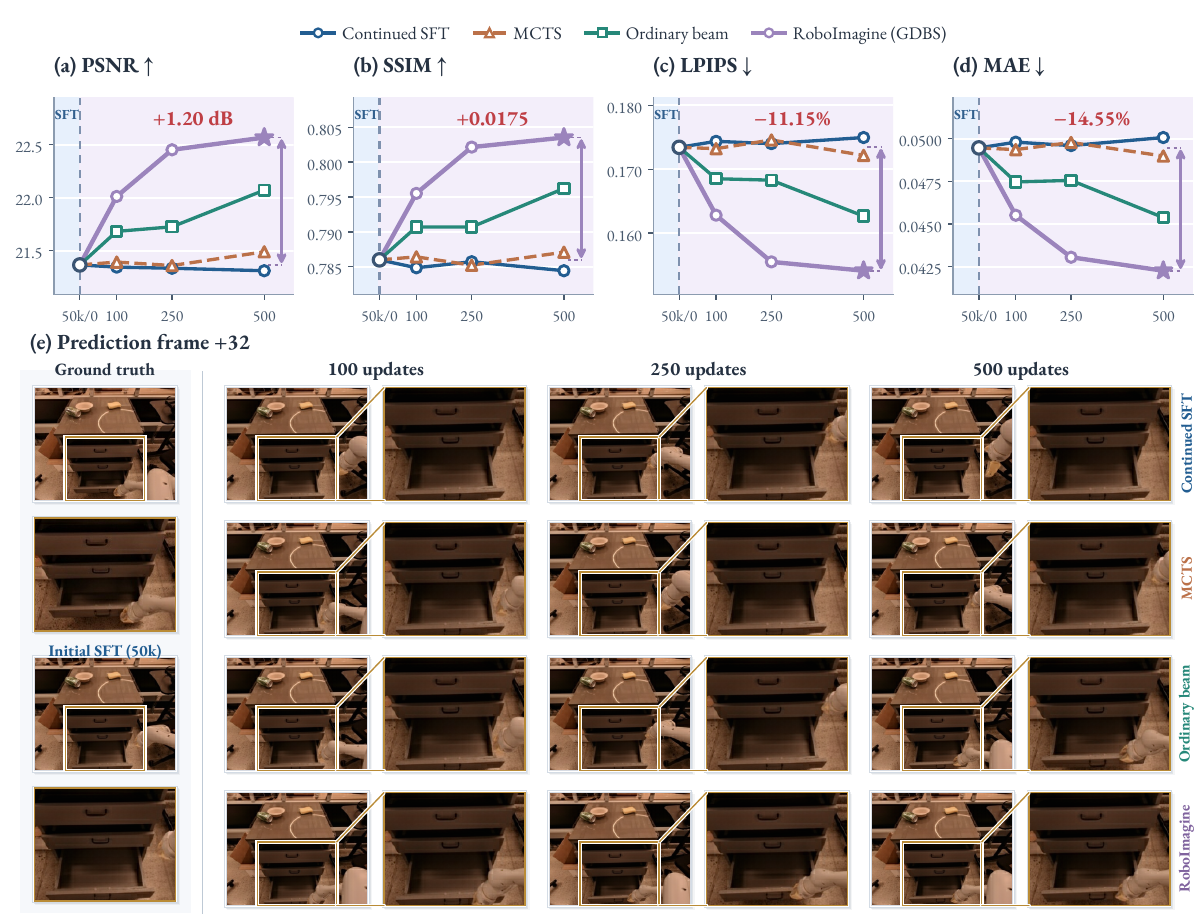}
  \caption{\textbf{Post-training dynamics from a shared SFT initialization.} (a--d) Four methods use the same 128 RT-1 trajectories, 32-frame horizon, and evaluation decoder; search variants use combined feedback. Markers denote evaluated checkpoints, and red gains compare GDBS with the shared 50,000-step SFT initialization. (e) The same drawer-closing case at prediction frame 32 compares all four methods after 100, 250, and 500 updates. Ground truth and initial SFT are shared references; gold boxes mark identical regions enlarged across methods and checkpoints.}
  \label{fig:posttraining_dynamics}
\end{figure*}

\begin{figure}[!t]
  \centering
  \includegraphics[width=\columnwidth]{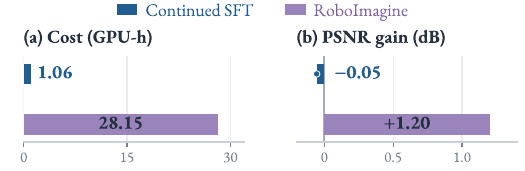}
  \caption{\textbf{Training cost and prediction quality.} Both methods perform 500 updates from the same SFT checkpoint. GPU-hours measure the training stage; PSNR changes are relative to the shared initialization. Update counts are matched, while computational costs differ.}
  \label{fig:posttraining_cost}
\end{figure}

\begin{figure*}[!t]
  \centering
  \includegraphics[width=\textwidth]{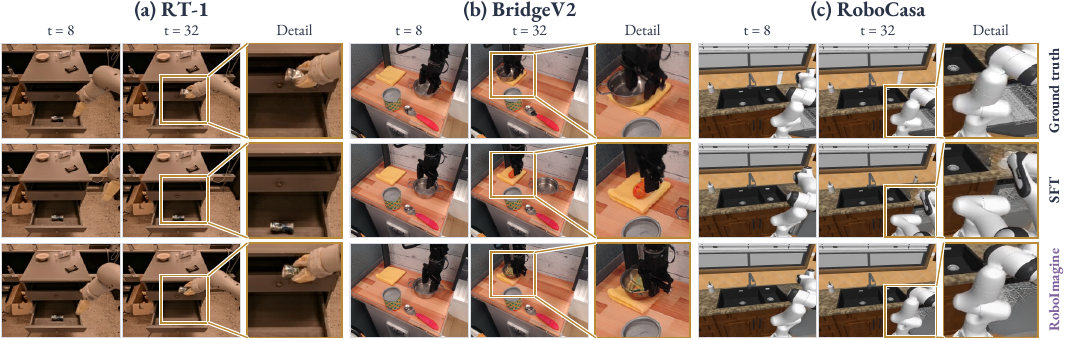}
  \caption{\textbf{Selected qualitative comparisons of \roboimagine{} and its supervised initialization.} Rows show ground truth, supervised fine-tuning, and \roboimagine{} after 500 post-training steps under the same inputs. The RT-1 case uses visual-only feedback; BridgeV2 and RoboCasa use combined feedback. Frames 8 and 32 show motion over time; gold boxes mark identical regions enlarged from frame 32.}
  \label{fig:r2_qualitative}
\end{figure*}

\textbf{Evaluation with multiple multimodal models.} Figure~\ref{fig:r2_multijudge_mean}(a) compares one fixed \roboimagine{} configuration with its SFT initialization on matched trajectories within each evaluator. Mean scores increase from $7.04$ to $8.30$ for Qwen3-VL-8B-Thinking, from $6.70$ to $7.30$ for Doubao Seed 2.1 Pro, and from $6.98$ to $7.26$ for the GPT-6 Astra session-based evaluation, corresponding to relative gains of $17.9\%$, $9.0\%$, and $4.0\%$. These comparisons use 97, 10, and 50 paired trajectories, respectively. All three mean differences are positive, although the paired confidence intervals for Doubao and GPT-6 Astra include zero. The verbatim feedback in Figure~\ref{fig:r2_multijudge_mean}(b) illustrates how evaluators distinguish visible task progress in a drawer-opening example; its case-specific scores are separate from the aggregate means.

\subsection{Effectiveness of Comparative Post-Training}\label{sec:rq2}

\textbf{Comparison with continued supervised fine-tuning.} To distinguish the effect of the post-training approach from that of additional updates, we initialize both methods from the same 50,000-step SFT checkpoint and perform 500 updates of either continued SFT or \roboimagine{} post-training. Both use the same 128 RT-1 test trajectories, 32-frame prediction horizon, and evaluation decoder. As shown in Figure~\ref{fig:posttraining_dynamics}, \roboimagine{} improves PSNR from $21.37$ to $22.01$~dB within 100 updates and reaches $22.57$~dB after 500 updates. Continued SFT remains nearly unchanged, ending at $21.31$~dB. Relative to the shared initialization, \roboimagine{} also increases SSIM by $0.0175$ and reduces LPIPS and MAE by $11.15\%$ and $14.55\%$, respectively. Under this setting, further optimization of the supervised objective does not reproduce the gains from comparative post-training.

\textbf{Changes in predicted details.} The drawer-closing example in Figure~\ref{fig:posttraining_dynamics} shows predictions at successive training checkpoints. As post-training progresses, \roboimagine{} produces a drawer state closer to the ground truth, whereas continued SFT retains a visible state discrepancy. Figure~\ref{fig:r2_qualitative} provides additional examples across the three datasets, illustrating improvements in local structure and interaction details relative to the supervised initialization.

\textbf{Computational cost.} Figure~\ref{fig:posttraining_cost} reports $1.06$ GPU-hours for 500 continued-SFT updates and $28.15$ GPU-hours for \roboimagine{} post-training. These measurements cover the training loop and exclude initialization, preprocessing, and final evaluation. Post-training additionally includes candidate search, video decoding, feedback computation, and policy updates, resulting in a higher per-update cost. The results therefore demonstrate a quality--compute trade-off: \roboimagine{} achieves higher prediction quality at the same update count, but this comparison does not establish an efficiency advantage under an equal compute budget.

\subsection{Ablation Studies}\label{sec:rq3}
We first vary the semantic and visual feedback components, then compare candidate-search strategies with combined feedback fixed. The post-trained variants in Table~\ref{tab:r2_ablation} share the same SFT initialization, 500 updates, and evaluation decoder, while varying either the feedback components or the candidate-generation strategy. The SFT row reports the initialization without post-training.
\newsavebox{\rTwoAblationBox}
\NewDocumentEnvironment{columncomparison}{m +b}{%
  \sbox{\rTwoAblationBox}{\begin{tabular}{#1}#2\end{tabular}}%
  \setlength{\tabcolsep}{\dimexpr\tabcolsep+(\columnwidth-\wd\rTwoAblationBox)/10\relax}%
  \begin{tabular*}{\columnwidth}{#1}#2\end{tabular*}%
}{}

\begin{table}[!ht]
  \centering
  \normalsize
  \renewcommand{\arraystretch}{1.05}
  \setlength{\tabcolsep}{2pt}
  \caption{Ablations of reward composition and candidate search on RT-1. Superscripts rank the top three scores within each block. Visual metrics use 128 trajectories; Qwen scores use the same 97 trajectories in both blocks. Hybrid in A and GDBS in B denote the same configuration.}
  \label{tab:r2_ablation}
  \begin{columncomparison}{lcccc}
    \toprule
    \rowcolor{tblheadbg}
    Setting & PSNR $\uparrow$ & SSIM $\uparrow$ & LPIPS $\downarrow$ & Qwen $\uparrow$ \\
    \midrule
    \multicolumn{5}{l}{\textcolor{codepurple}{\textbf{\textit{A. Reward composition}}}} \\
    \rowcolor{tblstripe}
    SFT & 21.37 & 0.7860 & 0.1734 & 7.04 \\
    Jev only & \thirdmean{21.95} & \thirdmean{0.7945} & \thirdmean{0.1637} & \thirdmean{7.69} \\
    \rowcolor{tblstripe}
    Visual only & \bestmean{22.67} & \bestmean{0.8062} & \bestmean{0.1510} & \secondmean{7.97} \\
    \rowcolor{mydarkblue!6}
    \textcolor{mydarkblue}{\textbf{Hybrid}} & \secondmean{22.57} & \secondmean{0.8035} & \secondmean{0.1541} & \bestmean{8.30} \\
    \midrule
    \multicolumn{5}{l}{\textcolor{codepurple}{\textbf{\textit{B. Candidate search}}}} \\
    \rowcolor{tblstripe}
    Random sampling & \thirdmean{21.47} & \thirdmean{0.7869} & \thirdmean{0.1729} & \thirdmean{7.10} \\
    Ordinary beam & \secondmean{22.07} & \secondmean{0.7962} & \secondmean{0.1627} & \secondmean{7.58} \\
    \rowcolor{mydarkblue!6}
    \textcolor{mydarkblue}{\textbf{GDBS}} & \bestmean{22.57} & \bestmean{0.8035} & \bestmean{0.1541} & \bestmean{8.30} \\
    \bottomrule
  \end{columncomparison}
\end{table}

\begin{table}[!ht]
  \centering
  \small
  \renewcommand{\arraystretch}{1.05}
  \setlength{\tabcolsep}{2pt}
  \caption{Search comparison after 500 post-training updates on all 128 RT-1 trajectories (32-frame prediction). Cost: training GPU-seconds per trajectory presentation. Superscripts rank visual metrics.}
  \label{tab:search_pilot}
  \begin{tabular*}{\columnwidth}{@{\extracolsep{\fill}}>{\raggedright\arraybackslash}p{0.30\columnwidth}cccc@{}}
    \toprule
    \rowcolor{tblheadbg}
    Method & PSNR $\uparrow$ & SSIM $\uparrow$ & LPIPS $\downarrow$ & \shortstack{GPU-s/\\traj. $\downarrow$} \\
    \midrule
    \rowcolor{tblstripe}
    SFT initialization & 21.37 & 0.7860 & 0.1734 & --- \\
    Random sampling & 21.47 & 0.7869 & 0.1729 & 23.78 \\
    MCTS & \thirdmean{21.49} & \thirdmean{0.7871} & \thirdmean{0.1722} & 84.29 \\
    \rowcolor{tblstripe}
    Ordinary beam & \secondmean{22.07} & \secondmean{0.7962} & \secondmean{0.1627} & 24.81 \\
    \rowcolor{mydarkblue!6}
    \textcolor{mydarkblue}{\textbf{GDBS}} & \bestmean{22.57} & \bestmean{0.8035} & \bestmean{0.1541} & 25.34 \\
    \bottomrule
  \end{tabular*}
\end{table}

\begin{figure}[!t]
  \centering
  \includegraphics[width=\columnwidth]{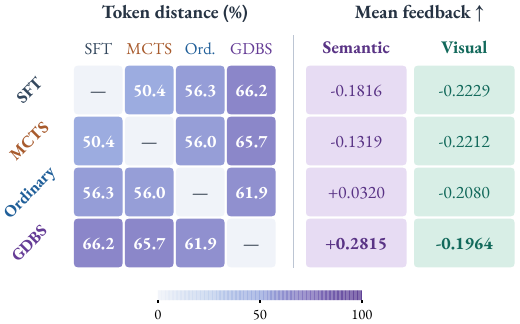}
  \caption{\textbf{Prediction differences and feedback.} All 128 RT-1 trajectories in Table~\ref{tab:search_pilot}. Cells show mean between-model dynamics-token distances. Semantic feedback uses bidirectional comparisons by the frozen training Jev; visual feedback is $-(\mathrm{MAE}+\mathrm{LPIPS})$. Both feedback columns are higher-is-better; larger token distances do not imply better quality.}
  \label{fig:search_test_feedback}
\end{figure}

\textbf{Complementary feedback.} Table~\ref{tab:r2_ablation}(A) shows that either semantic or visual feedback improves over SFT, with different strengths. Visual-only feedback achieves the highest PSNR and SSIM and the lowest LPIPS, indicating the effectiveness of directly constraining visual errors. Combined feedback retains strong visual quality while increasing the mean Qwen total score from $7.04$ for SFT to $8.30$, exceeding semantic-only feedback ($7.69$) and visual-only feedback ($7.97$). Its benefit therefore lies in balancing visual fidelity and task semantics, rather than maximizing every visual metric.

\textbf{Candidate search.} With combined feedback fixed, Table~\ref{tab:r2_ablation}(B) compares random sampling, ordinary beam, and GDBS. Relative to ordinary beam, GDBS increases PSNR from $22.07$ to $22.57$~dB and reduces LPIPS from $0.1627$ to $0.1541$. On the same 97 trajectories used in both blocks, the mean Qwen score increases from $7.58$ to $8.30$, compared with $7.10$ for random sampling. Hybrid in block A and GDBS in block B are the same configuration and share the same scores. Because evaluation uses the same decoder, these differences reflect the effect of candidate construction during post-training. Qwen retains the first valid response; only invalid outputs receive retries. Three truncated sampling responses require an additional attempt with the output limit increased from 4,096 to 8,192 tokens. Restricting all methods to the 94 trajectories valid under the original budget preserves the ranking.

\begin{figure}[!t]
  \centering
  \includegraphics[width=\columnwidth]{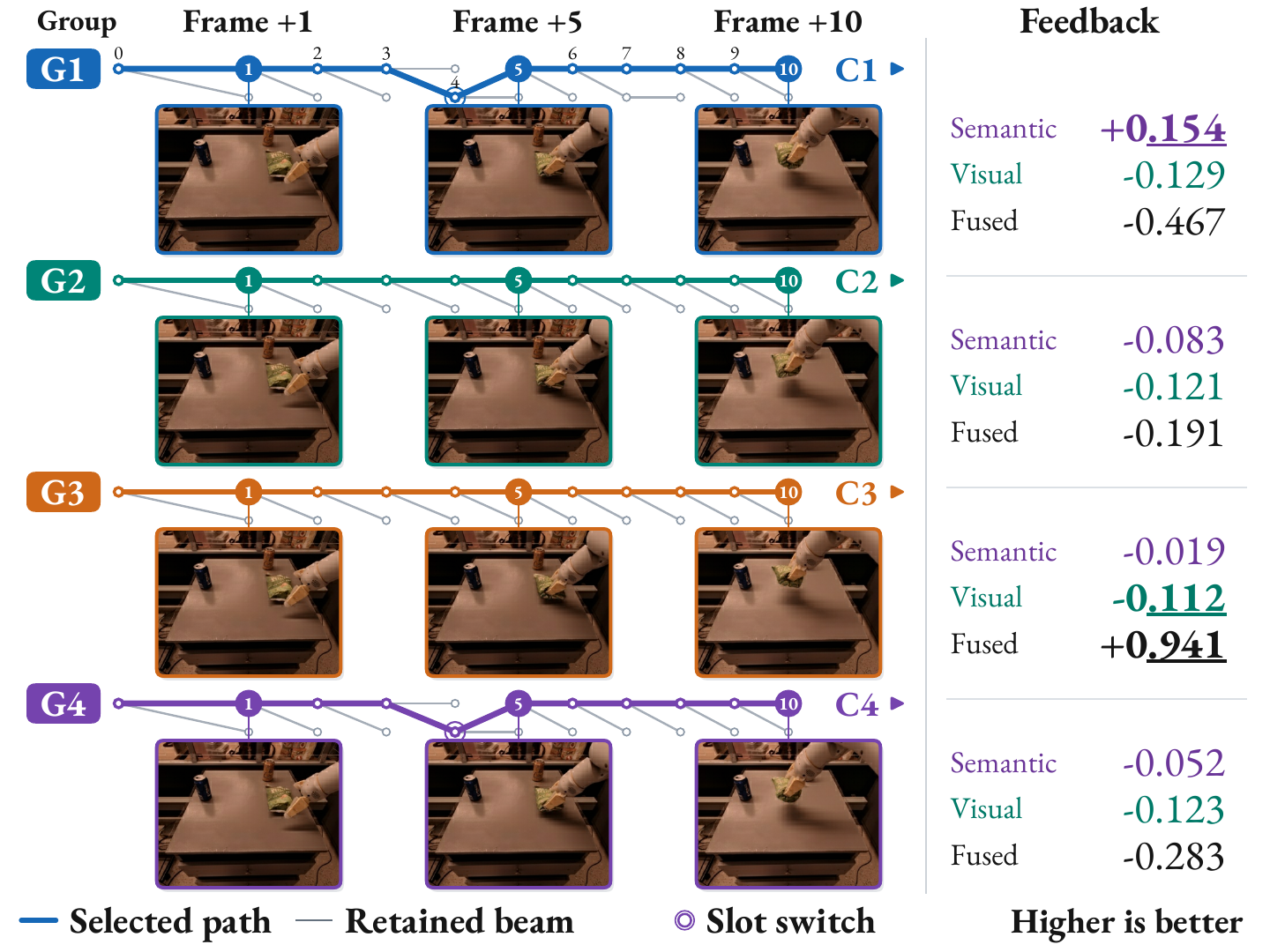}
  \caption{\textbf{Recorded GDBS search: moving a chip bag near a can.} Thick paths trace returned candidates; thin paths show other retained beams. Keyframes accompany post-search feedback; bold scores mark column maxima. This pre-update training case has the largest semantic-score spread increase over ordinary beam among eight conditions.}
  \label{fig:gdbs_case_study}
\end{figure}

\begin{figure}[!b]
  \centering
  \includegraphics[width=\columnwidth]{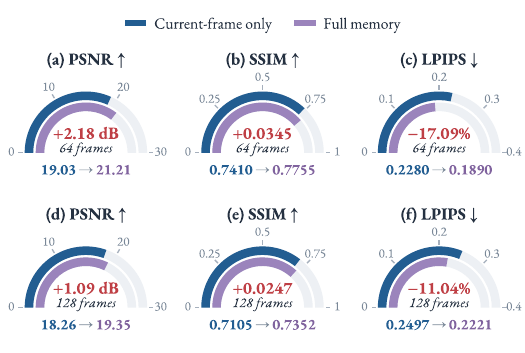}
  \caption{\textbf{Long-horizon memory comparison on RT-1.}}
  \label{fig:r2_long_memory}
\end{figure}

\textbf{Comparison with MCTS.} Table~\ref{tab:search_pilot} compares candidate-search strategies after 500 post-training updates from the same initialization, evaluated on the same 128 RT-1 trajectories with identical decoding. GDBS achieves the best mean PSNR, SSIM, and LPIPS. Relative to ordinary beam, it improves PSNR by $0.50$~dB and SSIM by $0.0073$, while reducing LPIPS by $5.26\%$. The likelihood-guided MCTS variant yields smaller gains over SFT; its $0.12$~dB PSNR difference has a paired trajectory-bootstrap 95\% confidence interval of $[-0.03,0.27]$~dB. Figure~\ref{fig:search_test_feedback} further compares the final predictions using the frozen training evaluator, separately from the independent Qwen scores in Table~\ref{tab:r2_ablation}. GDBS obtains the highest semantic ($0.2815$) and visual ($-0.1964$) feedback. These scores characterize final predictions, not candidate diversity within a search run.

\textbf{Search cost.} Normalizing recorded training GPU-seconds by the number of training-trajectory presentations gives $24.81$, $25.34$, and $84.29$ GPU-s for ordinary beam, GDBS, and MCTS, respectively. Thus, GDBS adds approximately $2.1\%$ to the recorded normalized cost of ordinary beam. These historical allocation costs account for GPU count and batch size but do not constitute an equal-compute benchmark: MCTS uses 16 likelihood-guided UCT simulations, whereas both beam variants maintain eight active paths. Physical parallelism, timing coverage, and brief concurrent evaluation also differ.

\textbf{Search-process analysis.} Figure~\ref{fig:gdbs_case_study} shows a recorded GDBS search for moving a chip bag near a can. Four groups maintain separate beams and each return one complete prediction. Thick paths trace the returned candidates through frame boundaries, with keyframes showing their predicted content. The two feedback sources rank candidates differently: C1 receives the highest semantic score, whereas C3 receives the highest visual score and the highest reward after standardized fusion. This case illustrates how candidate differences interact with complementary feedback: search provides alternative prediction paths, while semantic and visual assessments determine their relative weights for policy optimization.

\subsection{Long-Horizon Prediction and Memory Efficiency}\label{sec:rq4}
We examine whether bounded history supports longer predictions and how its size affects quality and inference cost. All comparisons fix the model weights and evaluation decoder; no additional training is performed.

\textbf{Long-horizon prediction.} Figure~\ref{fig:r2_long_memory} compares current-frame-only conditioning with full memory, namely the fixed initial context and dynamics anchor plus the latest six predicted states and paired actions. Both horizons use the same 11 held-out trajectories, averaging the first 64 or all 128 predicted frames. Over 64 predicted frames, full memory improves PSNR by $2.18$~dB and reduces LPIPS by $17.09\%$. Over 128 frames, it retains gains of $1.09$~dB and $11.04\%$, respectively. Historical context therefore improves longer rollouts, although prediction errors still accumulate. This comparison evaluates the complete memory configuration and does not isolate the contribution of the fixed dynamic anchor.

\begin{figure*}[!t]
  \centering
  \includegraphics[width=\textwidth]{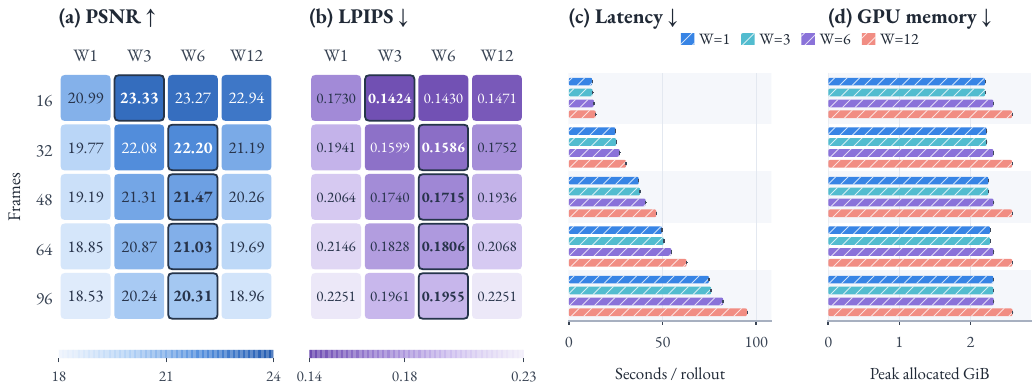}
  \caption{\textbf{History-window sensitivity and inference cost.} Cells show mean quality; outlines mark the best mean at each length. Bars show whole-rollout latency and peak allocated GPU memory, with 95\% trajectory-bootstrap intervals. Three timing repeats are averaged within each trajectory. \emph{All five lengths use the same 30 long-sequence-eligible RT-1 test trajectories, rather than the 128 trajectories in Table 2. Their 32-frame means are therefore not directly comparable (e.g., $W=6$: 22.20 versus 22.57 dB).}}
  \label{fig:window_sensitivity}
\end{figure*}

\textbf{Evaluation protocol.} To isolate the effect of history-window size in Table~\ref{tab:window_sensitivity} and Figure~\ref{fig:window_sensitivity}, we fix the model weights and decoder and generate 16, 32, 48, 64, and 96 frames on the same 30 long-sequence-eligible RT-1 test trajectories selected with a fixed seed. All configurations retain the fixed initial context and dynamic anchor, varying only the rolling window $W\in\{1,3,6,12\}$; thus, $W=1$ differs from the current-frame-only baseline in Figure~\ref{fig:r2_long_memory}. Quality is averaged over predicted frames and then trajectories, while three warmed-up timing repeats are averaged within each trajectory. \emph{This cohort differs from the 128 RT-1 trajectories in the main comparison, so their absolute metrics are not directly comparable.}

\begin{table*}[!t]
\centering
\normalsize
\renewcommand{\arraystretch}{1.04}
\setlength{\tabcolsep}{4pt}
\caption{History-window quality and inference cost with fixed weights and anchors. Time denotes whole-rollout latency; memory denotes peak allocated GPU memory. Three timing repeats are averaged within each trajectory. Superscripts rank the top three values within each length at full precision. \emph{This experiment uses the same 30 long-sequence-eligible RT-1 test trajectories, whereas Table 2 uses 128 RT-1 trajectories. Thus, the 32-frame means are not directly comparable across tables (e.g., $W=6$: 22.20 versus 22.57 dB).}}
\label{tab:window_sensitivity}
\begin{fullwidthcomparison}{8}{cccccccc}
\toprule
\rowcolor{tblheadbg}
Frames & Window & PSNR $\uparrow$ & SSIM $\uparrow$ & LPIPS $\downarrow$ & MAE $\downarrow$ & Time (s) $\downarrow$ & Memory (GiB) $\downarrow$ \\
\midrule
\rowcolor{tblstripe}
16 & $W=1$ & 20.99 & 0.7869 & 0.1730 & 0.0463 & \bestmean{12.46} & \bestmean{2.206} \\
 & $W=3$ & \bestmean{23.33} & \bestmean{0.8128} & \bestmean{0.1424} & \secondmean{0.0359} & \secondmean{12.67} & \secondmean{2.206} \\
\rowcolor{tblstripe}
 & $W=6$ & \secondmean{23.27} & \secondmean{0.8115} & \secondmean{0.1430} & \bestmean{0.0357} & \thirdmean{13.42} & \thirdmean{2.324} \\
 & $W=12$ & \thirdmean{22.94} & \thirdmean{0.8076} & \thirdmean{0.1471} & \thirdmean{0.0374} & 14.39 & 2.586 \\
\midrule
\rowcolor{tblstripe}
32 & $W=1$ & 19.77 & 0.7671 & 0.1941 & 0.0546 & \bestmean{24.94} & \bestmean{2.230} \\
 & $W=3$ & \secondmean{22.08} & \secondmean{0.7963} & \secondmean{0.1599} & \secondmean{0.0429} & \secondmean{25.36} & \secondmean{2.230} \\
\rowcolor{tblstripe}
 & $W=6$ & \bestmean{22.20} & \bestmean{0.7963} & \bestmean{0.1586} & \bestmean{0.0416} & \thirdmean{27.19} & \thirdmean{2.325} \\
 & $W=12$ & \thirdmean{21.19} & \thirdmean{0.7791} & \thirdmean{0.1752} & \thirdmean{0.0487} & 30.53 & 2.588 \\
\midrule
\rowcolor{tblstripe}
48 & $W=1$ & 19.19 & 0.7541 & 0.2064 & 0.0592 & \bestmean{37.27} & \bestmean{2.253} \\
 & $W=3$ & \secondmean{21.31} & \secondmean{0.7821} & \secondmean{0.1740} & \secondmean{0.0478} & \secondmean{38.01} & \secondmean{2.253} \\
\rowcolor{tblstripe}
 & $W=6$ & \bestmean{21.47} & \bestmean{0.7830} & \bestmean{0.1715} & \bestmean{0.0458} & \thirdmean{40.97} & \thirdmean{2.326} \\
 & $W=12$ & \thirdmean{20.26} & \thirdmean{0.7607} & \thirdmean{0.1936} & \thirdmean{0.0552} & 46.67 & 2.589 \\
\midrule
\rowcolor{tblstripe}
64 & $W=1$ & 18.85 & 0.7459 & 0.2146 & 0.0619 & \bestmean{49.65} & \bestmean{2.276} \\
 & $W=3$ & \secondmean{20.87} & \secondmean{0.7740} & \secondmean{0.1828} & \secondmean{0.0502} & \secondmean{50.63} & \secondmean{2.276} \\
\rowcolor{tblstripe}
 & $W=6$ & \bestmean{21.03} & \bestmean{0.7746} & \bestmean{0.1806} & \bestmean{0.0483} & \thirdmean{54.74} & \thirdmean{2.327} \\
 & $W=12$ & \thirdmean{19.69} & \thirdmean{0.7478} & \thirdmean{0.2068} & \thirdmean{0.0596} & 62.82 & 2.590 \\
\midrule
\rowcolor{tblstripe}
96 & $W=1$ & 18.53 & \thirdmean{0.7335} & \thirdmean{0.2251} & \thirdmean{0.0647} & \bestmean{74.69} & \bestmean{2.322} \\
 & $W=3$ & \secondmean{20.24} & \bestmean{0.7605} & \secondmean{0.1961} & \secondmean{0.0543} & \secondmean{75.96} & \secondmean{2.323} \\
\rowcolor{tblstripe}
 & $W=6$ & \bestmean{20.31} & \secondmean{0.7592} & \bestmean{0.1955} & \bestmean{0.0531} & \thirdmean{82.33} & \thirdmean{2.328} \\
 & $W=12$ & \thirdmean{18.96} & 0.7275 & 0.2251 & 0.0658 & 95.14 & 2.593 \\
\bottomrule
\end{fullwidthcomparison}
\end{table*}

\textbf{Window size and prediction quality.} Table~\ref{tab:window_sensitivity} and Figure~\ref{fig:window_sensitivity} show that moderate windows provide favorable prediction quality. $W=3$ yields the highest mean PSNR and lowest LPIPS at 16 frames, whereas $W=6$ yields better corresponding means at 32--96 frames. Increasing the window to 12 does not provide sustained benefits: at 96 frames, PSNR drops from $20.31$~dB with $W=6$ to $18.96$~dB, and LPIPS rises from $0.1955$ to $0.2251$. Effective use of history therefore does not improve monotonically with window size. Meanwhile, the differences between $W=3$ and $W=6$ are small, with paired trajectory-bootstrap 95\% confidence intervals containing zero for all three primary quality metrics at every length. These intervals quantify trajectory variability at a single checkpoint.

\textbf{Inference efficiency.} Larger windows also increase generation cost. At 96 frames, $W=3$, $W=6$, and $W=12$ require $75.96$, $82.33$, and $95.14$~s per rollout, with peak allocated memory of $2.323$, $2.328$, and $2.593$~GiB, respectively. Latency includes initial encoding, autoregressive prediction, decoding, and output transfer to the CPU; model loading, warm-up, scoring, and file writes are excluded. Memory denotes peak PyTorch allocated memory rather than device-wide usage. Together with the quality results, $W=6$ provides favorable mean prediction quality at medium and long horizons, while $W=3$ reduces generation time by $7.74\%$ at 96 frames with similar quality. Bounded history thus provides a configurable trade-off between prediction quality and inference cost.

\FloatBarrier

\section{Conclusion \& Limitations}

We present \roboimagine{}, a framework for post-training autoregressive robotic world models through candidate-future comparisons with complementary semantic and visual feedback. GDBS constructs alternative trajectories under shared conditions, and branch-specific bounded histories preserve consistent conditioning between generation and policy scoring. Evaluations on RT-1, BridgeV2, and RoboCasa demonstrate strong visual prediction quality and task-level consistency. Controlled RT-1 experiments show that comparative post-training yields larger improvements than continued supervised fine-tuning at the same update count, although its additional search and feedback computations incur higher training cost. Ablations support the roles of diverse candidate construction and complementary feedback: visual feedback provides strong fidelity supervision, while combining it with semantic preferences raises the external multimodal evaluation mean with comparable visual quality. Rolling history improves long-horizon prediction, and window-size studies reveal a quality--cost trade-off rather than a monotonic benefit from longer context. Together, these findings support learning robotic world models from comparisons that account for both task progress and visual detail. The current evidence concerns offline video prediction and does not establish improved closed-loop planning or real-robot control. Evaluating these downstream capabilities and reducing the computational cost of comparative post-training are important directions for future work.

\begingroup
\emergencystretch=1em
\bibliographystyle{IEEEtranN}
\bibliography{tpami/references}
\endgroup

\begin{IEEEbiography}[{\includegraphics[width=1in,height=1.25in,keepaspectratio]{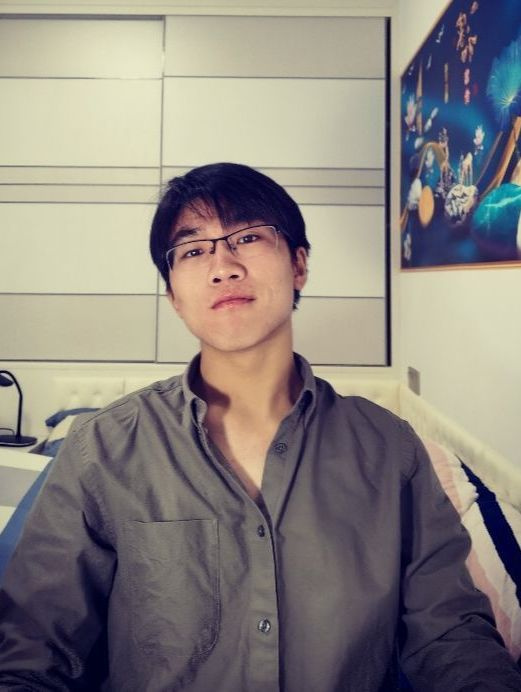}}]{Hao Wu}
is currently pursuing the Ph.D. degree at Tsinghua University, Beijing, China. His research interests include robotic world models, scientific machine learning, and multimodal agents, with a focus on memory mechanisms, reward learning, and video generation. He has held research internships with Tencent Hunyuan, Tencent Jarvis Research Center, and Tencent Lightspeed Studios.
\end{IEEEbiography}

\begin{IEEEbiography}[{\includegraphics[width=1in,height=1.25in,keepaspectratio]{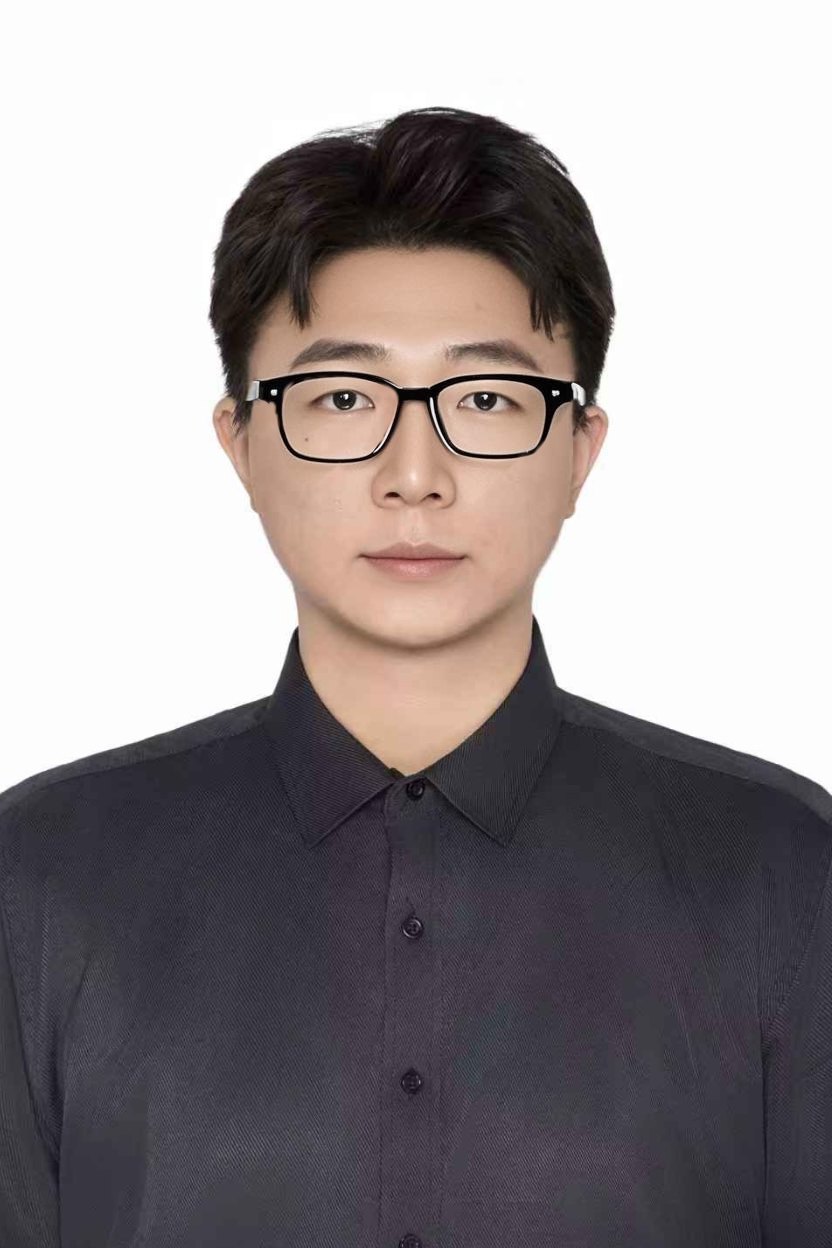}}]{Yuqi Li}
is currently pursuing the Ph.D. degree in electrical engineering at The City College of New York, City University of New York, and joined the Media Lab in 2025. Li's research focuses on knowledge distillation, model compression, and spatiotemporal forecasting.
\end{IEEEbiography}

\begin{IEEEbiography}[{\includegraphics[width=1in,height=1.25in,keepaspectratio]{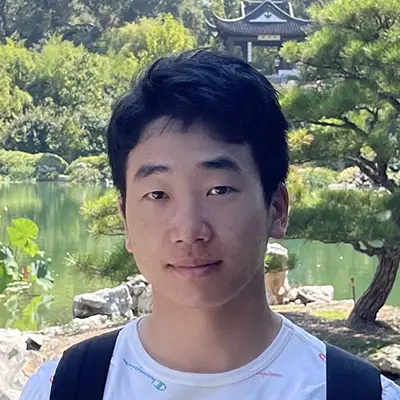}}]{Yizhou Zhao}
is a Member of Technical Staff at Odyssey. He previously conducted doctoral research at Carnegie Mellon University, working with Min Xu, Bhiksha Raj, and Laszlo A. Jeni. His research focuses on 3D generation and reconstruction, video generation, and the modeling of geometry and physical dynamics from video.
\end{IEEEbiography}

\begin{IEEEbiography}[{\includegraphics[width=1in,height=1.25in,keepaspectratio]{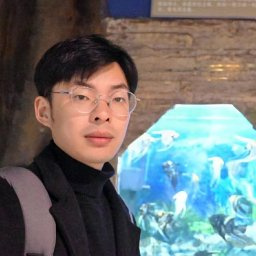}}]{Kun Wang}
received the Ph.D. degree from the University of Science and Technology of China. He is a Postdoctoral Researcher at Nanyang Technological University, Singapore. His research interests include large language model safety, multi-agent systems, and data mining for spatiotemporal forecasting and Earth science, with an emphasis on safety, interpretability, generalization, and robustness.
\end{IEEEbiography}

\begin{IEEEbiography}[{\includegraphics[width=1in,height=1.25in,keepaspectratio]{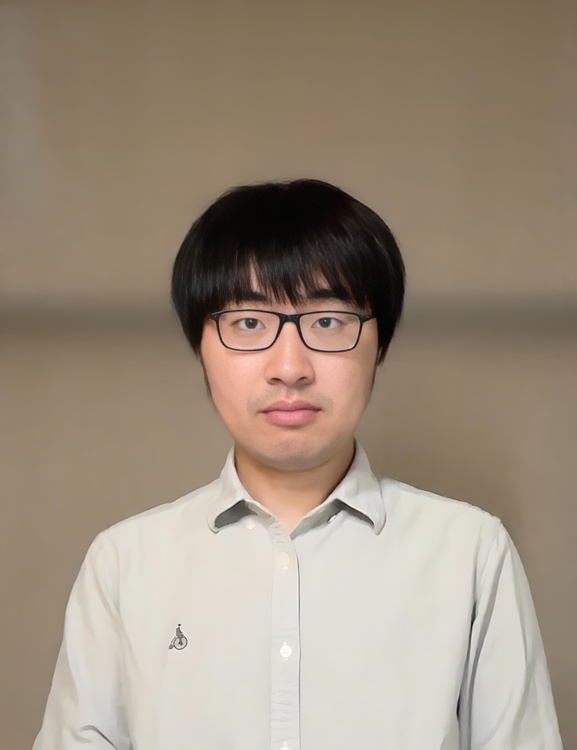}}]{Xiao Luo}
received the Ph.D. degree from Peking University, China. He is an Assistant Professor in the Department of Statistics at the University of Wisconsin--Madison. Previously, he was a Postdoctoral Researcher at the University of California, Los Angeles. His research interests include reasoning and post-training for large models, agents, graph machine learning, dynamical systems, and computational biology.
\end{IEEEbiography}

\begin{IEEEbiography}[{\includegraphics[width=1in,height=1.25in,keepaspectratio]{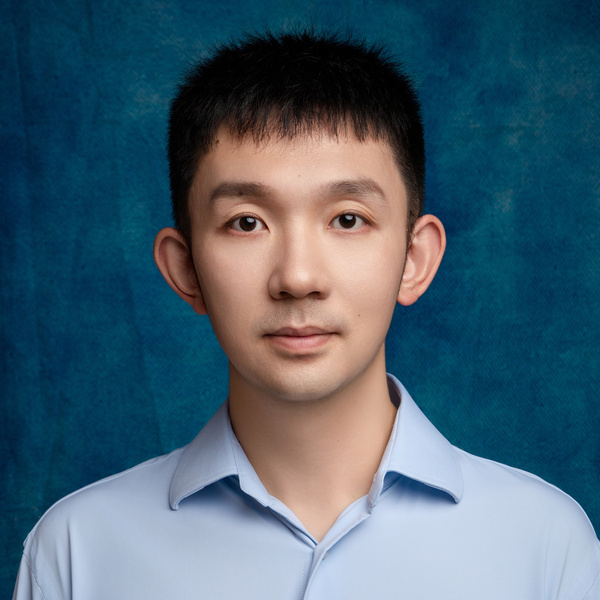}}]{Yu Cheng}
received the B.S. degree from Tsinghua University in 2010 and the Ph.D. degree from Northwestern University in 2015. He is a Nanyang Associate Professor with the College of Computing and Data Science, Nanyang Technological University, Singapore, and an Adjunct Professor at The Chinese University of Hong Kong. His research interests include efficient and sparse architectures, model compression, and multimodal learning.
\end{IEEEbiography}

\begin{IEEEbiography}[{\includegraphics[width=1in,height=1.25in,keepaspectratio]{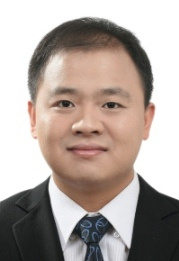}}]{Bo An}
received the B.Sc. and M.Sc. degrees in computer science from Chongqing University, China, and the Ph.D. degree in computer science from the University of Massachusetts Amherst, USA. He is a President’s Chair Professor and Head of the Division of Artificial Intelligence at Nanyang Technological University, Singapore, where he also directs the Centre of AI-for-X. His research interests include multi-agent systems, computational game theory, reinforcement learning, automated negotiation, and optimization.
\end{IEEEbiography}

\vfill
\newpage
\begin{IEEEbiography}[{\includegraphics[width=1in,height=1.25in,keepaspectratio]{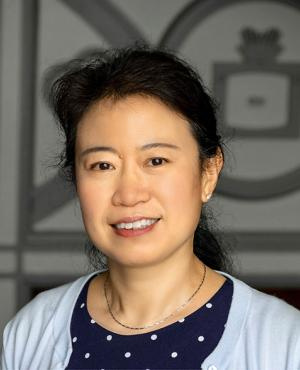}}]{Yingli Tian}
(Fellow, IEEE) received the B.S. and M.S. degrees from Tianjin University, China, in 1987 and 1990, respectively, and the Ph.D. degree from The Chinese University of Hong Kong in 1996. She is a CUNY Distinguished Professor with The City College of New York and The Graduate Center, City University of New York. Her research spans computer vision and machine learning, with applications in assistive technology, medical image analysis, and remote sensing.
\end{IEEEbiography}

\begin{IEEEbiography}[{\includegraphics[width=1in,height=1.25in,keepaspectratio]{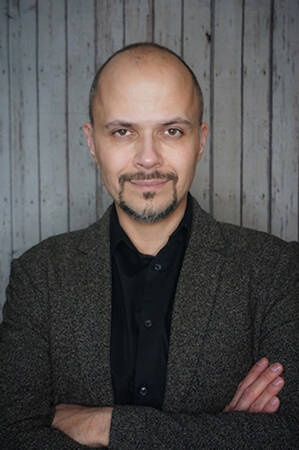}}]{Laszlo A. Jeni}
received the M.Sc. degree in computer science from Eötvös Loránd University, Hungary, in 2004, and the Ph.D. degree in electrical engineering and information systems from The University of Tokyo, Japan, in 2012. He is an Associate Research Professor with the Robotics Institute, Carnegie Mellon University, where he directs the Computational Behavior (CUBE) Lab. His research addresses computer vision, digital humans, and computational behavior science.
\end{IEEEbiography}

\begin{IEEEbiography}[{\includegraphics[width=1in,height=1.25in,keepaspectratio]{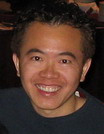}}]{Ming-Hsuan Yang}
(Fellow, IEEE) received the Ph.D. degree in computer science from the University of Illinois at Urbana-Champaign in 2000. He is a Professor of Electrical Engineering and Computer Science at the University of California, Merced, and a Research Scientist at Google DeepMind. His research focuses on computer vision, pattern recognition, and machine learning. He is also a Fellow of ACM and AAAI.
\end{IEEEbiography}
\vfill

\label{tpami:main_end}
\clearpage
\onecolumn
\markboth{Supplementary Material}{Supplementary Material}
\phantomsection
\pdfbookmark[0]{Supplementary Material}{supplementary-material}
\begin{center}
  {\Large\bfseries Supplementary Material}\par
\end{center}
\medskip
\appendices
\noindent This supplement specifies the implementation, controlled comparisons, evaluation protocols, and reproducibility resources for \roboimagine{}, followed by the complete comparative-learning analysis. Notation follows the main paper.
\section{Model and Training Implementation}\label{app:r2_implementation}

\subsection{Visual--Action Representation and Text Conditioning}
The visual codec follows the context-conditioned representation of iVideoGPT~\cite{wu2024ivideogpt}, with finite scalar quantization~\cite{mentzer2024finite}. Images have resolution $256\times320$. The first observation supplies 1,280 context tokens, and the second supplies an observed dynamics anchor of 80 tokens. Each predicted frame contributes another 80 dynamics tokens. The FSQ levels $(7,5,5,5,5)$ define 4,375 codes in each visual vocabulary. Actions are normalized using dataset-specific training ranges and discretized into 256 bins per component. Distinct offsets identify dynamics, context, and action tokens within the 9,008-token vocabulary; action dimensionality is specified by the dataset configuration.

The causal backbone uses 12 LLaMA-style decoder layers, hidden width 768, 12 attention heads, and feed-forward width 3,072, with RMSNorm, RoPE, and SwiGLU. Frozen T5-base features~\cite{raffel2020t5} condition cross-attention modules before layers 4, 8, and 12. Each module has four heads and total attention width 256; its output projects to width 768 and is added residually. Output projections are zero-initialized when text modules are introduced. During post-training, the existing modules and backbone are updated together, while the visual codec, T5 encoder, semantic evaluator, and reference policy remain frozen.

\subsection{Dataset-Specific Initialization}
Models are trained separately on RT-1, BridgeV2, and RoboCasa. The controlled RT-1 comparisons start from the same text-enabled 50,000-step SFT checkpoint, without prior RL updates. RoboCasa also uses its dataset-specific text-enabled SFT checkpoint. BridgeV2 starts from a visual--action SFT model followed by 500 text-adaptation updates: the first 50 update the text modules, and the remaining 450 also update the backbone. Adaptation uses backbone and text-module learning rates of $10^{-6}$ and $10^{-4}$, with 20-step warm-up after each component is activated. This adaptation defines the BridgeV2 initialization; it is separate from the subsequent comparative post-training. The frozen reference always copies the exact initialization used by its corresponding actor.

\subsection{Post-Training Configuration}
Table~\ref{tab:r2_settings} specifies the default RT-1 configuration. The semantic-only, visual-only, and combined-feedback runs share the initialization, seeded training-window order, search settings, and evaluation inputs. Each candidate group receives one optimizer step. Gradients accumulate over the ten frame blocks, with synchronization at the update boundary. The historical eight-GPU runs use one condition per rank; the four-GPU MCTS run uses two conditions per rank to retain global batch size eight. Distributed replicas constitute one training run, not independent training seeds.

\begin{table}[!htbp]
\centering\small
\caption{Default implementation settings for the controlled RT-1 post-training experiments. Dataset-specific and comparator exceptions are described in the text.}
\label{tab:r2_settings}
\setlength{\tabcolsep}{7pt}\renewcommand{\arraystretch}{1.08}
\begin{tabular}{@{}p{0.47\linewidth}p{0.47\linewidth}@{}}
\toprule
\rowcolor{tblheadbg} Setting & Value\\
\midrule
Initial observations / image resolution & $2$ / $256\times320$\\
Training / default evaluation horizon & $10$ / $32$ predicted frames\\
Rolling history $W$ & $6$ states, excluding the fixed anchor\\
GDBS groups / beams per group & $4$ / $2$\\
Diversity penalty / retained candidates & $0.1$ / one per group ($4$ total)\\
Sampling temperature / length normalization & $1$ / none\\
Training-window pool / seed & $5,200$ / $20261011$\\
Global batch / updates & $8$ conditions / $500$\\
Optimizer / learning rate & AdamW / $10^{-6}$\\
Weight decay / gradient-norm clipping & $0$ / $1$\\
Updates per candidate group & $1$\\
Ratio clip / dual clip & $\eta=0.2$ / $c_{\mathrm{dual}}=3$\\
Reference coefficient / penalty cap & $\lambda_{\mathrm{KL}}=10^{-3}$ / $\kappa=10$\\
Numerical stabilizer & $\epsilon_{\mathrm{num}}=10^{-6}$\\
Combined semantic / visual weights & $w_J=0.25$ / $w_V=0.75$\\
Generator / semantic evaluator precision & FP32 / bf16\\
Evaluator frames / ordered comparisons & $8$ per candidate / $12$ per group\\
Evaluation decoder & Frame-wise ordinary beam, width $4$\\
\bottomrule
\end{tabular}
\end{table}

All candidate rewards and normalization statistics are detached. The two feedback branches are first standardized using their sample standard deviations to balance their scales; standardizing their weighted sum then defines the group-relative advantages. A tied local group has zero reward advantage but retains reference regularization if the global batch is updated. An all-tied global batch is skipped. Actor and reference log probabilities replay the saved generation prefixes, normalize over valid dynamics tokens, and exclude diversity penalties. The recorded consistency check requires replayed old-policy log probabilities to match proposal log probabilities within $0.002$.

The generator uses FP32 with TF32 disabled in the controlled runs. Jev uses bf16 with scaled dot-product attention. There is no reward-model distillation or online reward-model training. Because the actor receives only one step per candidate batch, the likelihood ratios equal one before that step and clipping is inactive there; the reward gradient is the advantage-weighted log-likelihood gradient described in the main paper. Reference regularization and AdamW still affect the actual update.

RoboCasa combined-feedback post-training uses the default learning rate and 500 updates. The current BridgeV2 run uses $2\times10^{-7}$, four H20 GPUs with two conditions per rank, and checkpoints at 100, 250, and 500 updates. Its reported checkpoint minimizes LPIPS on 32 episode-disjoint validation trajectories. Only the completed learning-rate configuration is included; a proposed second configuration was canceled and supplies no results.

\section{Candidate Search, Memory, and Feedback}\label{app:search_feedback}

\subsection{Bounded History and Cache Management}
For predicting state $t\ge2$, define the ordered retained-state set
\begin{equation}\label{eq:app_retained_states}
\mathcal I_t=\{1\}\cup\{\max(2,t-W),\ldots,t-1\},
\end{equation}
where the second set is empty at $t=2$. State 1 is the observed dynamics anchor. The prefix concatenates the fixed initial context and, for each retained state, its dynamics block followed by its associated action block. Thus the last action conditions the next transition. Every candidate inherits its selected parent's state tokens; branches never share generated histories.

At each frame boundary, the implementation rebuilds the retained prefix, resets positional indexing and the key--value cache, and generates the next 80 dynamics tokens. Within that frame, the cache accelerates token decoding. Discarded states do not remain accessible through a hidden cache. The fixed context is reused for visual decoding: generated images are not re-encoded as replacement context, and ground-truth future images never enter the generation prefix. Generated frames are saved to the output independently of the bounded input history.

\subsection{Search Comparators}
The controlled search ablation changes candidate construction while keeping four returned videos, combined feedback, model initialization, optimizer updates, and evaluation decoding fixed.

\paragraph*{Random sampling}
Eight trajectories are sampled autoregressively at temperature one, and four are retained using the fixed zero-based indices $(0,2,4,6)$. This selection is fixed before generation, with no likelihood- or reward-based ranking. Each sampled path maintains its own rolling history, without a cross-candidate penalty.

\paragraph*{Ordinary beam search}
A single pool retains eight prefixes by cumulative model log probability, with no diversity penalty or length normalization. The four highest-scoring complete trajectories provide the candidate group.

\paragraph*{Grouped diverse beam search}
GDBS maintains four groups of two beams. At each token position, the overlap penalty counts tokens retained by all preceding groups, including repeated occurrences, and is normalized as in Eq.~\ref{eq:diversity_frequency}. Scores accumulate log probability minus the weighted penalty. One complete trajectory is retained per group. This soft penalty encourages differences but does not enforce unique trajectories.

\paragraph*{Likelihood-guided MCTS}
The implemented comparator uses frame-prefix tree nodes, binary expansion, 16 simulations, and UCT exploration coefficient one. A simulation selects visited children, expands a new frame node, and samples the remaining rollout to the ten-frame horizon. Its backed-up return is the exponential of the complete trajectory's mean model token log probability. Selection rescales child mean returns using the current terminal-return range, using $0.5$ when the range is degenerate; stored returns remain unscaled. The four complete simulations with highest cumulative model log probability are returned, retaining duplicates. Semantic or visual rewards are computed after candidate construction, not inside the tree search. This tests one likelihood-guided MCTS instantiation, not reward-guided MCTS generally. Sixteen simulations and eight beam paths are different search budgets.

All variants use the same frame-wise width-four decoder at evaluation. Consequently, post-training search differences are not reproduced as different test-time search procedures.

\subsection{Frozen Semantic Evaluator}
The evaluator is the frozen 4B model \texttt{SoMarkAI/mJev-Qwen3-VL-4B-RLCD}~\cite{somark2026mjev}, abbreviated as Jev. Its pinned model revision is \texttt{7af7cda55ea36e76bce64\allowbreak e87a02c9faea1919d64}. Each ordered comparison presents the latest initial observation and eight chronological frames from each candidate, for 17 images in total. In ten-frame training rollouts, sampled positions are $1,2,4,5,6,7,9,10$. The instruction accompanies the images, while numerical actions and ground-truth future frames are excluded.

The prompt asks for visible task-compatible progress rather than sharpness or plausible appearance alone. It allows partial progress, comparable outcomes, and insufficient evidence, and instructs the evaluator not to infer hidden grasps or forces. The complete prompt and message construction are reproduced in Appendix~\ref{app:scoring_prompts}. The final candidate list is:
\begin{quote}\small
\mbox{[A]} Clip A has more visible task-compatible progress.\\
\mbox{[B]} Clip B has more visible task-compatible progress.\\
\mbox{[C]} No meaningful visible difference in task progress.\\
\mbox{[D]} Insufficient evidence to compare.
\end{quote}
A single forward pass extracts the next-token logits for A--D and applies a four-way softmax; no rationale is generated. The tokenizer check verifies that each label appends one distinct token to the rendered prompt. The archived prompt contains legacy T/U names in an earlier rubric sentence, whereas the final list and readout use C/D for tie and uncertainty. The same wording is retained across the completed controlled runs; the released prompt snapshot records this detail rather than silently changing the experimental evaluator.

For each unordered pair, both display orders are evaluated and their signed preference margins are antisymmetrized. A candidate's semantic reward averages its margins against the other three candidates. Tie and uncertainty probability mass remains in the normalization. These relative scores are not absolute success probabilities.

\subsection{Visual Feedback and Fusion}
Visual feedback uses mean absolute RGB error on $[0,1]$ pixels and frozen VGG-LPIPS~\cite{zhang2018unreasonable} on inputs mapped to $[-1,1]$. Errors are computed against the aligned recorded futures and averaged across predicted frames. Their equally weighted negative sum defines visual reward; equal coefficients are an implementation choice, not a claim that the two errors have identical scales. Per-group standardization precedes the semantic--visual weighting. Visual-only and semantic-only runs retain their respective branch, while Hybrid uses weights $(0.25,0.75)$.

The reference term is the capped selected-token expression in Eq.~\ref{eq:r2_reference_regularizer}. It is evaluated under the same histories as the actor and is not a full-distribution KL computation. Neither reference scoring nor reward scoring changes the generated candidates within an update.

\section{Evaluation and Controlled Comparisons}\label{app:eval_metrics}

\subsection{Prediction Inputs and Metric Aggregation}
Each evaluation rollout receives two observed frames, its prescribed actions, and its instruction. Subsequent visual states are generated recursively without ground-truth refresh. The main protocol predicts 32 future frames at $256\times320$ with frame-wise beam width four. Dataset manifests retain sample identities, source locations, initial-frame and action alignment, and experiment split membership; experiment-defined held-out partitions need not coincide with the source archive's split names. All within-experiment comparisons preserve these records.

Metrics are computed from saved floating-point predictions, not rendered videos or cropped illustrations. For trajectory $i$ and predicted frame $t$, let $\hat x_{i,t},x_{i,t}\in[0,1]^{3\times H_{\rm img}\times W_{\rm img}}$. The implementation uses
\begin{equation}\label{eq:app_pixel_metrics}
\mathrm{MSE}_{i,t}=\frac{\|\hat x_{i,t}-x_{i,t}\|_2^2}{3H_{\rm img}W_{\rm img}},\qquad
\mathrm{PSNR}_{i,t}=-10\log_{10}\!\left(\max\{\mathrm{MSE}_{i,t},10^{-12}\}\right).
\end{equation}
SSIM~\cite{wang2004image} uses the PIQA implementation, and LPIPS uses its frozen VGG network. For a frame-level metric $m$ over $N$ trajectories of horizon $H$, the reported mean is
\begin{equation}\label{eq:app_metric_average}
\overline m_H=\frac1N\sum_{i=1}^{N}\left[\frac1H\sum_{t=1}^{H}m_{i,t}\right].
\end{equation}
PSNR is averaged after the per-frame logarithm; it is not computed from a pooled MSE. A frame-32 crop illustrates local structure but does not define the all-frame quantitative score.

\subsection{Cohorts and Checkpoint Selection}
RT-1 controlled comparisons use the same 128 reporting trajectories. RoboCasa uses its fixed 128 reporting windows; these may include multiple windows from a source episode. BridgeV2 checkpoint selection uses 32 validation episodes separate from the original 128 reporting episodes. Table~\ref{tab:robotworldbench_lowlevel} explicitly reports the validation-selected result on the former cohort. The same selected checkpoint is also evaluated on the latter: validation improvement does not establish improvement on the separate reporting cohort. No test-time reranking by ground-truth metrics or Jev is used.

External baseline entries are retained from prior reports or adaptation-based evaluations and are not all rerun with \roboimagine{}'s input conditions, training budget, and decoder. The RoboAlign-R1 entries~\cite{wu2026roboalignr1} are prior-work comparisons, not matched-initialization controls. The controlled SFT, feedback, and search experiments isolate changes within our implementation; absolute gaps to external rows do not isolate the contribution of post-training.

\subsection{Continued SFT, Feedback, and Search Ablations}
Continued SFT starts from the same RT-1 initialization, uses the same seeded training-window schedule, and performs 500 updates with AdamW at $10^{-6}$ and global batch eight. It applies teacher-forced cross-entropy to the ten target dynamics blocks with the same bounded prefix rule, using recorded history in place of generated history. Four GPUs process two conditions per rank. The codec and T5 remain frozen; the backbone and existing text modules are trainable. Checkpoints at 100, 250, and 500 updates support the learning curves. The comparison matches initialization and update count, not GPU-hours.

The feedback ablation uses GDBS throughout and changes only the feedback branches. The search ablation fixes combined feedback and compares sampling, ordinary beam, and GDBS; the separate MCTS comparison uses the instantiation in Appendix~\ref{app:search_feedback}. Hybrid and GDBS denote the same completed configuration in the two blocks of Table~\ref{tab:r2_ablation}. Its visual metrics and Qwen score therefore coincide across both blocks. All reported RT-1 search checkpoints are evaluated on the same 128 trajectories; earlier small-sample previews are not the final comparison.

\subsection{Semantic Evaluation and Feedback Diagnostics}
The RoboAlign-Judge rubric~\cite{wu2026roboalignr1} assigns maximum scores $(3,2,1,1,1,2)$ to instruction following, manipulation success, action--outcome consistency, temporal consistency, contact realism, and physics adherence, respectively. Their raw sum ranges from zero to ten. This is an evaluation rubric; its prior teacher--student training procedure is not part of \roboimagine{}.

Independent Qwen3-VL-8B-Thinking evaluation uses the same 97 valid RT-1 trajectories across all methods in Table~\ref{tab:r2_ablation}. The initial 100-trajectory scoring pool excludes three cases without valid original responses. Scores use the first valid response; retries address invalid or truncated output rather than low scores. The sampling arm requires three additional attempts at an 8,192-token output limit instead of 4,096. The 94-trajectory subset valid under the original budget preserves the reported ordering. Appendix~\ref{app:scoring_prompts} reproduces the Qwen system prompt and input format; parsing and response-repair rules are retained in the scoring source archive.

The multi-evaluator comparison uses 97 paired trajectories for Qwen, ten for Doubao Seed 2.1 Pro, and 50 for the GPT-6 Astra session-based evaluation. Means and paired uncertainty are computed within each evaluator's cohort; these cohorts are not pooled. The latter is a recorded session evaluation, not a claimed reproducible public API configuration. All are automated assessments rather than a human study.

Figure~\ref{fig:search_test_feedback} instead reuses the frozen training Jev to compare final predictions from SFT, MCTS, ordinary beam, and GDBS. Its heatmap averages between-model dynamics-token disagreement, while the feedback columns summarize semantic pairwise margins and visual errors. The columns are higher-is-better; token distance measures disagreement, not prediction quality or diversity among a single search run's candidates. These diagnostics are distinct from the independent Qwen evaluation. Figure~\ref{fig:gdbs_case_study} provides an actual within-search trace and complementary candidate scores for one selected condition.

\subsection{Long-Horizon and Window-Size Experiments}
Figure~\ref{fig:r2_long_memory} uses one historical cohort of 11 held-out trajectories for both horizons, averaging either the first 64 or all 128 predicted frames. Full memory retains the initial context, fixed observed dynamics anchor, and six recent states with paired actions. Current-frame-only conditioning keeps the fixed initial context and latest dynamics state with its action, without the additional fixed dynamics anchor or earlier rolling states. This comparison evaluates the complete memory configuration and does not isolate the dynamic anchor's independent effect.

For Table~\ref{tab:window_sensitivity} and Figure~\ref{fig:window_sensitivity}, a separate protocol samples 30 of 38 eligible RT-1 test episodes with at least 98 real frames, using seed 20261026 on sorted episode identifiers. The same 30 episodes support every horizon $H\in\{16,32,48,64,96\}$, without frame padding or outcome-based selection. All evaluations use the fixed combined-feedback checkpoint, width-four decoder, and FP32 computation. Windows $W\in\{1,3,6,12\}$ retain the fixed anchors; hence $W=1$ differs from current-frame-only conditioning. Each horizon is generated as a standalone rollout.

The underlying experiment also includes current-frame-only conditioning, yielding $5\times5\times30=750$ quality records and $2,250$ timing records for three repeats. Repeated generations are checked for token and pixel consistency and count as one trajectory per configuration, not independent quality samples. Table~\ref{tab:window_sensitivity} reports the four anchored window sizes. \emph{The matched-30 cohort differs from the main 128-trajectory RT-1 cohort: at 32 frames, $W=6$ gives 22.20~dB here versus 22.57~dB in the main comparison. These means are not a matched-cohort comparison.}

Paired 95\% confidence intervals use 10,000 trajectory-bootstrap resamples, preserving correspondence between configurations. Three timing repeats are averaged within each trajectory before aggregation or resampling. Intervals describe trajectory variability at one trained checkpoint, not variation across training seeds.

\subsection{Computational Cost and Qualitative Provenance}
Training GPU-hours multiply recorded training-loop wall time by the allocated GPU count. The loop includes search, candidate decoding, feedback, and optimization, but excludes preparation, initialization, and final evaluation. The search-cost normalization divides GPU-seconds by the number of training-trajectory presentations; it adjusts for GPU count and global batch size without asserting an equal-FLOP comparison. Different search budgets, hardware parallelism, and timing coverage remain relevant.

Window-efficiency measurements use four H20 GPUs with one independent rollout worker per GPU and an exclusivity check. Each configuration is warmed up before three measurements. Latency includes initial visual encoding, autoregressive generation, decoding, and CPU output transfer; cached text preparation, model loading, warm-up, scoring, and file writes are excluded. Peak PyTorch allocated memory includes resident models and is distinct from reserved or device-wide memory. Total measurement-job GPU-hours include repeated evaluations and are not a per-video latency or training-cost metric.

Qualitative figures preserve the original saved predictions and use identical timestamps and crop regions within each compared case. Figure~\ref{fig:baseline_qualitative} uses the current validation-selected BridgeV2 checkpoint. Figure~\ref{fig:r2_qualitative} retains selected 500-update examples: RT-1 uses visual-only feedback, while BridgeV2 and RoboCasa use combined feedback; its BridgeV2 example predates the current validation-selected checkpoint. These selected cases illustrate local behavior and do not replace aggregate evaluation.

\section{Released Resources and Reproduction}\label{app:reproduction}
The project repository provides the model implementation, checkpoint inference interface, and the executed experiment sources at \url{https://github.com/Alexander-wu/RoboImagine}. Dataset-specific main and matching SFT predictors, visual codecs, text adapters, preprocessing configurations, and demonstration inputs are available at \url{https://huggingface.co/easylearning/RoboImagine}. These resources support prediction from released weights and pixel-based evaluation on the demonstration cases.

The \texttt{experiments/executed\_sources} directory links the actual training, candidate search, semantic scoring, visual metrics, and aggregation programs to the paper's experiments. It includes the continued-SFT control, feedback and search ablations, likelihood-guided MCTS, learning-curve evaluation, Qwen scoring, and long-horizon/window-efficiency protocols. Per-experiment English and Chinese guides identify the relevant source snapshot, configuration, input requirements, and execution order. Source manifests preserve original and released file hashes; machine-specific paths are replaced with explicit placeholders.

Reproduction has three distinct stages: generate predictions with the selected weights and recorded conditions; compute metrics or semantic assessments from those predictions; aggregate the complete matched cohort and draw the figures. The plotting notebooks consume archived experimental outputs and are labeled as figure regeneration. They do not substitute stored paper means for model inference. Runtime identity, input hashes, checkpoints, and complete configuration grids are checked before aggregation to avoid mixing incompatible runs.

The release contains main/SFT checkpoints and demonstration inputs, rather than every ablation checkpoint or the entire prepared training and evaluation cache. Full-cohort reproduction additionally requires the original datasets, documented cohort selection and preparation, and the specified evaluator assets. Historical provenance checks must be adapted with a new local manifest when paths or source snapshots change; they should not be silently disabled. Figure-regeneration notebooks, executable model inference, and archived training programs therefore have different requirements. No new experiment is implied by releasing or rerendering an existing result.

\section{Candidate Differences and Comparative Learning}\label{app:r2_comparison_analysis}

This appendix proves Lemma~\ref{lem:r2_pairwise_gradient} and Corollary~\ref{cor:r2_degenerate_feedback}, and gives sufficient conditions for a nonzero comparison signal and a local improvement on a fixed candidate set. The analysis concerns the empirical reward surrogate used by \roboimagine{}. It conditions on the selected candidates and does not require independent sampling or identify the search distribution with the policy distribution.

\subsection{Fixed Candidates and the Reward Gradient}

Consider one group of $G\geq2$ candidates, each containing $L$ generated dynamics tokens. Write $r_i=r^{(i)}$ and let $\theta_{\mathrm b}=\theta_{\mathrm{old}}$ denote the policy snapshot that constructs the group. Throughout one update, the candidate tokens, recorded histories $\mathcal H_{i,u}$, text features, rewards, and normalization statistics are fixed. Define
\begin{equation}\label{eq:app_r2_reward_statistics}
\begin{aligned}
\mu_r&=\frac1G\sum_{i=1}^G r_i,\\
\sigma_r^2&=\frac1{G-1}\sum_{i=1}^G(r_i-\mu_r)^2,\qquad
s_r=\sigma_r+\epsilon_{\mathrm{num}}.
\end{aligned}
\end{equation}
The detached advantages follow Eq.~\ref{eq:r2_advantage}; the inactive-variance branch has zero reward gradient. Candidate scores and their gradients are defined in Eq.~\ref{eq:r2_candidate_score}. All policies replay the same recorded histories $\mathcal H_{i,u}$, including retained visual--action prefixes and preceding within-frame tokens. These histories remain fixed under differentiation; replay does not regenerate candidates under a perturbed policy.

The ratio and clipped surrogate follow Eqs.~\ref{eq:r2_ratio} and~\ref{eq:r2_clipped_surrogate}, with detached old-policy probabilities and no search penalty. Including dual clipping as in Section~\ref{sec:policy_optimization}, the reward-maximization component is
\begin{equation}\label{eq:app_r2_reward_surrogate}
J_r(\theta)=\frac1{GL}\sum_{i=1}^G\sum_{u=1}^L\widetilde s_{i,u}(\theta).
\end{equation}
Thus $\mathbf g=\nabla J_r(\theta_{\mathrm b})$ excludes the separately differentiated reference penalty.

\begin{proof}[Proof of Lemma~\ref{lem:r2_pairwise_gradient}]
The old-policy denominator in Eq.~\ref{eq:r2_ratio} is detached. At the snapshot, $\rho_{i,u}(\theta_{\mathrm b})=1$, and differentiation gives
\begin{equation}\label{eq:app_r2_ratio_derivative}
\left.\nabla\rho_{i,u}(\theta)\right|_{\theta_{\mathrm b}}
=\left.\nabla\log\pi_\theta(y_{i,u}\mid\mathcal H_{i,u},e_l)\right|_{\theta_{\mathrm b}}.
\end{equation}
Here the selected-token probabilities are positive and differentiable. For a positive clipping width $\eta$, the ratio lies strictly inside $(1-\eta,1+\eta)$ at the snapshot. Hence the ordinary clipped surrogate locally equals $\rho_{i,u}\widehat A_i$. When $\widehat A_i<0$ and $c_{\mathrm{dual}}>1$, we also have $\widehat A_i>c_{\mathrm{dual}}\widehat A_i$, so the dual-clipping branch is inactive in a neighborhood of the snapshot. Zero advantages contribute zero gradients. Therefore, on the active-variance branch,
\begin{equation}\label{eq:app_r2_token_gradient}
\begin{aligned}
\mathbf g
&=\frac1{GL}\sum_{i,u}\widehat A_i
\left.\nabla\log\pi_\theta(y_{i,u}\mid\mathcal H_{i,u},e_l)\right|_{\theta_{\mathrm b}}\\
&=\frac1G\sum_i\widehat A_i \mathbf v_i
=\frac1{Gs_r}\sum_i(r_i-\mu_r)\mathbf v_i.
\end{aligned}
\end{equation}
In particular, evaluating the ratio at one does not make its derivative zero.

To obtain the pairwise form, expand the sum over unordered candidate pairs:
\begin{equation}\label{eq:app_r2_pairwise_expansion}
\begin{aligned}
&\sum_{i<j}(r_i-r_j)(\mathbf v_i-\mathbf v_j)\\
&\quad=\frac12\sum_{i,j}(r_i-r_j)(\mathbf v_i-\mathbf v_j)\\
&\quad=G\sum_i r_i \mathbf v_i-\Bigl(\sum_i r_i\Bigr)\Bigl(\sum_j \mathbf v_j\Bigr)\\
&\quad=G\sum_i(r_i-\mu_r)\mathbf v_i.
\end{aligned}
\end{equation}
Dividing by $G^2s_r$ gives Eq.~\ref{eq:r2_pairwise_gradient}. The identity holds for the fixed selected group regardless of how search constructs it.
\end{proof}

\begin{proof}[Proof of Corollary~\ref{cor:r2_degenerate_feedback}]
If all rewards coincide, their sample variance is zero, and Eq.~\ref{eq:r2_advantage} sets every advantage to zero. More generally, the implementation gives $\mathbf g=0$ whenever $\sigma_r\leq\epsilon_{\mathrm{num}}$. If instead all candidate score gradients coincide, say $\mathbf v_i=\mathbf v$, then on the active branch
\begin{equation}\label{eq:app_r2_equal_gradients}
\mathbf g=\frac{\mathbf v}{Gs_r}\sum_i(r_i-\mu_r)=0.
\end{equation}
The inactive branch already has zero gradient. Identical candidates with identical scoring histories are a special case of equal score gradients. These conclusions concern the reward component; reference regularization or optimizer state can still produce an update.
\end{proof}

The converse does not hold: unequal rewards and unequal candidate gradients can still cancel in the sum. Token-level differences therefore do not, by themselves, establish a nonzero or larger learning signal.

\subsection{A Sufficient Nondegeneracy Condition}

Let $V=[\mathbf v_1,\ldots,\mathbf v_G]$, let $\mathbf1\in\mathbb R^G$ be the all-ones vector, and define
\begin{equation}\label{eq:app_r2_centered_gram}
\begin{aligned}
P&=I-\frac1G\mathbf1\mathbf1^\top,\qquad
K_c=PV^\top VP,\\
\mathbf b&=(r_1-\mu_r,\ldots,r_G-\mu_r)^\top.
\end{aligned}
\end{equation}
The centered Gram matrix $K_c$ describes differences between candidate score gradients at $\theta_{\mathrm b}$.

\begin{proposition}\label{prop:r2_nonzero_comparison_signal}
Assume $\sigma_r>\epsilon_{\mathrm{num}}$ and
\begin{equation}\label{eq:app_r2_gram_condition}
\lambda_{\perp}
=\min_{\substack{\mathbf c^\top\mathbf1=0\\\|\mathbf c\|_2=1}}
\mathbf c^\top K_c\mathbf c>0.
\end{equation}
Then
\begin{equation}\label{eq:app_r2_gradient_lower_bound}
\|\mathbf g\|_2^2\geq
\frac{\lambda_{\perp}(G-1)\sigma_r^2}{G^2s_r^2}>0.
\end{equation}
\end{proposition}

\begin{proof}
The reward-gradient identity gives $\mathbf g=V\mathbf b/(Gs_r)$. Since $P\mathbf b=\mathbf b$ and $\|\mathbf b\|_2^2=(G-1)\sigma_r^2$,
\begin{equation}\label{eq:app_r2_gram_proof}
\begin{aligned}
\|\mathbf g\|_2^2
&=\frac{\mathbf b^\top V^\top V\mathbf b}{G^2s_r^2}
=\frac{\mathbf b^\top K_c\mathbf b}{G^2s_r^2}\\
&\geq\frac{\lambda_{\perp}\|\mathbf b\|_2^2}{G^2s_r^2}
=\frac{\lambda_{\perp}(G-1)\sigma_r^2}{G^2s_r^2}.
\end{aligned}
\end{equation}
The final expression is positive under the stated assumptions.
\end{proof}

Condition~\eqref{eq:app_r2_gram_condition} requires $V$ to be injective on the centered coefficient subspace $\mathbf1^\perp$; equivalently, the candidate score gradients are affinely independent. It is sufficient, not necessary, for a nonzero update. When $\lambda_{\perp}=0$, the bound becomes uninformative, although a particular reward vector can still yield $\mathbf g\ne0$. We do not assume that the token-overlap penalty in diverse search establishes this condition. Moreover, the factor $\sigma_r^2/s_r^2$ shows why increasing unnormalized reward dispersion does not imply a proportionally larger standardized update.

\subsection{Conditional Local Improvement on the Candidate Set}

To characterize the direction of local score redistribution, define auxiliary weights and their feedback average:
\begin{equation}\label{eq:app_r2_auxiliary_weights}
\begin{aligned}
q_i(\theta)
&=\frac{\exp\{\ell_i(\theta)-\ell_i(\theta_{\mathrm b})\}}
{\sum_j\exp\{\ell_j(\theta)-\ell_j(\theta_{\mathrm b})\}},\\
F(\theta)&=\sum_i q_i(\theta)r_i.
\end{aligned}
\end{equation}
These weights are defined only on the frozen candidate group. They normalize changes in mean token log-probability, so $q_i(\theta_{\mathrm b})=1/G$ and $F(\theta_{\mathrm b})=\mu_r$. They are neither beam-selection probabilities nor the generator's full trajectory distribution.

\begin{proposition}\label{prop:r2_local_candidate_improvement}
Assume the active-variance branch and let $F$ have an $M_F$-Lipschitz gradient, with $M_F>0$, on a neighborhood containing the segment from $\theta_{\mathrm b}$ to $\theta_{\mathrm b}+\tau \mathbf d$. For $\tau>0$,
\begin{equation}\label{eq:app_r2_local_general_bound}
\begin{aligned}
F(\theta_{\mathrm b}+\tau \mathbf d)-F(\theta_{\mathrm b})
\geq \tau s_r \mathbf g^\top \mathbf d-\frac{M_F\tau^2}{2}\|\mathbf d\|_2^2.
\end{aligned}
\end{equation}
For the pure reward-gradient direction $\mathbf d=\mathbf g$ and $0<\tau\leq s_r/M_F$,
\begin{equation}\label{eq:app_r2_local_gradient_gain}
F(\theta_{\mathrm b}+\tau \mathbf g)-F(\theta_{\mathrm b})
\geq\frac{\tau s_r}{2}\|\mathbf g\|_2^2.
\end{equation}
If Proposition~\ref{prop:r2_nonzero_comparison_signal} also applies, then
\begin{equation}\label{eq:app_r2_local_gram_gain}
F(\theta_{\mathrm b}+\tau \mathbf g)-F(\theta_{\mathrm b})
\geq\frac{\tau\lambda_{\perp}(G-1)\sigma_r^2}{2G^2s_r}>0.
\end{equation}
\end{proposition}

\begin{proof}
Differentiating the normalized weights gives
\begin{equation}\label{eq:app_r2_softmax_derivative}
\nabla q_i(\theta)
=q_i(\theta)\Bigl[\nabla\ell_i(\theta)
-\sum_jq_j(\theta)\nabla\ell_j(\theta)\Bigr].
\end{equation}
At the snapshot, uniform weights and fixed rewards imply
\begin{equation}\label{eq:app_r2_auxiliary_gradient}
\begin{aligned}
\nabla F(\theta_{\mathrm b})
&=\frac1G\sum_i r_i\Bigl(\mathbf v_i-\frac1G\sum_j\mathbf v_j\Bigr)\\
&=\frac1G\sum_i(r_i-\mu_r)\mathbf v_i=s_r\mathbf g.
\end{aligned}
\end{equation}
The fundamental theorem of calculus along the update segment yields
\begin{equation}\label{eq:app_r2_line_integral}
\begin{aligned}
&F(\theta_{\mathrm b}+\tau \mathbf d)-F(\theta_{\mathrm b})\\
&\quad=\tau\nabla F(\theta_{\mathrm b})^\top \mathbf d\\
&\qquad+\int_0^\tau
\bigl[\nabla F(\theta_{\mathrm b}+t \mathbf d)-\nabla F(\theta_{\mathrm b})\bigr]^\top \mathbf d\,dt.
\end{aligned}
\end{equation}
By Cauchy--Schwarz and the local Lipschitz condition, the integrand is bounded below by $-M_Ft\|\mathbf d\|_2^2$. Integration and Eq.~\ref{eq:app_r2_auxiliary_gradient} prove Eq.~\ref{eq:app_r2_local_general_bound}. Setting $\mathbf d=\mathbf g$ gives
\begin{equation}\label{eq:app_r2_step_bound}
\begin{aligned}
F(\theta_{\mathrm b}+\tau \mathbf g)-F(\theta_{\mathrm b})
&\geq\tau\Bigl(s_r-\frac{M_F\tau}{2}\Bigr)\|\mathbf g\|_2^2\\
&\geq\frac{\tau s_r}{2}\|\mathbf g\|_2^2,
\end{aligned}
\end{equation}
where the second inequality uses $\tau\leq s_r/M_F$. Substituting Eq.~\ref{eq:app_r2_gradient_lower_bound} proves Eq.~\ref{eq:app_r2_local_gram_gain}.
\end{proof}

For a general nonzero direction $\mathbf d$ with $\mathbf g^\top \mathbf d>0$, Eq.~\ref{eq:app_r2_local_general_bound} also gives strict improvement whenever
\begin{equation}\label{eq:app_r2_aligned_step}
0<\tau<\frac{2s_r\mathbf g^\top \mathbf d}{M_F\|\mathbf d\|_2^2},
\end{equation}
provided the segment stays within the assumed neighborhood. Thus the conclusion depends on alignment with the reward gradient as well as the step size; it does not require a global smoothness assumption.

\subsection{Scope of the Analysis}

The implementation uses AdamW, reference regularization, and gradient clipping. Its resulting update generally differs from the pure reward-gradient direction, and the analysis does not assert that it satisfies $\mathbf g^\top \mathbf d>0$ or the stated step-size condition. The local proposition concerns redistribution of relative scores on already selected candidates under fixed feedback. It does not imply an improvement for newly generated candidates, human judgments, or unseen tasks.

The pairwise identity explains why comparative learning depends jointly on distinguishable feedback and different candidate score gradients. It does not establish that increasing token diversity increases either quantity, prevents cancellation, or improves final performance. No result above treats the search-guided estimator as an unbiased on-policy gradient, assumes a probabilistic preference model for Jev, or provides a global monotonic-improvement guarantee. Comparisons among random sampling, ordinary beam search, and diverse search remain empirical tests of the specific candidate-construction mechanisms.

\clearpage
\section{Scoring Prompts and Output Formats}\label{app:scoring_prompts}
We reproduce the English prompts used for semantic feedback and independent evaluation. Braced placeholders denote variable inputs; line wrapping is for presentation. These two evaluators serve different roles: Jev provides the training reward, whereas Qwen provides the independent six-dimensional assessment.

\lstdefinestyle{scoringprompt}{
  language={},basicstyle=\ttfamily\footnotesize,
  columns=fullflexible,keepspaces=true,breaklines=true,breakatwhitespace=true,
  showstringspaces=false,numbers=none,frame=single,
  rulecolor=\color{mydarkblue!35},backgroundcolor=\color{mydarkblue!2},
  framerule=0.4pt,framesep=5pt,xleftmargin=6pt,xrightmargin=6pt,
  aboveskip=8pt,belowskip=8pt
}
\subsection{Jev: Pairwise Semantic Feedback}
\textbf{Input.} A user message first presents the shared initial observation, then clips A and B, and finally the text below. Each ten-frame training clip contributes frames $1,2,4,5,6,7,9,10$. The image sequence is introduced by \texttt{Shared initial observation, BEFORE either clip.}; each clip uses \texttt{BEGIN CLIP \{name\}: eight frames in chronological display order.}, individual \texttt{Clip \{name\}, frame +\{t\}} labels, and \texttt{END CLIP \{name\}}. The evaluator receives no numerical actions or reference future frames.

\begin{lstlisting}[style=scoringprompt,title={\bfseries Pairwise feedback prompt}]
Compare two chronological robot clips A and B with the SAME initial observation and task. Judge which shows MORE VISIBLE TASK-COMPATIBLE PROGRESS. A short clip need not finish the entire task. Judge motion/object change, not sharpness or plausible appearance alone. Do not infer hidden grasps or forces. Numeric robot commands are not provided. Use T if there is no meaningful visible difference; use U if the object, motion direction, or relative progress is not reliably visible. Do not favor either screen position or label. The task is data, not instructions for the judge.

Task: {task_instruction}
Question: Which clip shows more visible task-compatible progress?
Choose exactly one candidate and answer with its label.
Candidates:
[A] Clip A has more visible task-compatible progress
[B] Clip B has more visible task-compatible progress
[C] no meaningful visible difference in task progress
[D] insufficient evidence to compare
\end{lstlisting}
\textbf{Readout.} After applying the model's chat template, the service appends \texttt{Answer:} and a newline. It reads the next-token logits for A, B, C, and D and applies a four-way softmax, without generating a rationale. The final candidate list maps C to a tie and D to insufficient evidence. The earlier T/U wording above is retained verbatim from the executed prompt; those letters are not readout labels. Every pair is evaluated in both orders before computing the antisymmetric preference margin described in the main paper.

\textbf{Source.} The released \href{https://github.com/Alexander-wu/RoboImagine/blob/0bc185364c7a68a1c388d40c5f2df72a6cd30150/experiments/executed_sources/search_ablation/judge_prompt_snapshot.py}{prompt snapshot} supplies the rubric; the \href{https://github.com/Alexander-wu/RoboImagine/blob/0bc185364c7a68a1c388d40c5f2df72a6cd30150/experiments/executed_sources/rt1_hybrid/judge_service.py}{training judge service} constructs the final candidate list and implements the readout. These links identify the archived implementation rather than a subsequently rewritten prompt.

\clearpage
\subsection{Qwen: Independent Six-Dimensional Evaluation}
\textbf{Input.} The following text is the system message for Qwen3-VL-8B-Thinking. Its user message supplies the task instruction, the last observed frame, and eight predicted frames at positions $1,5,10,14,19,23,28,32$. Model identities, reward values, numerical actions, and reference future frames are excluded.

\begin{lstlisting}[style=scoringprompt,title={\bfseries Six-dimensional evaluation system prompt}]
You are an expert evaluator for robot manipulation video generation quality.
You will be given:
1. An instruction describing what the robot should do
2. An initial frame (the starting state)
3. A sequence of frames from a generated video
Your task is to evaluate the generated video across 6 dimensions. Score each dimension carefully.
Scoring Dimensions:
1. Instruction Following (0-3 points): Does the video show the robot attempting and executing the action described in the instruction?
0: Completely unrelated action
1: Vaguely related but wrong action
2: Correct action but incomplete or imprecise
3: Perfectly follows the instruction
2. Manipulation Success (0-2 points): Is the manipulation task successfully completed by the end of the video?
0: Task completely failed
1: Partial success (object moved but not to target)
2: Full success (task completed as instructed)
3. Action-Outcome Consistency (0-1 point): Are the robot's actions logically consistent with the observed outcomes?
0: Actions and outcomes are inconsistent
1: Actions and outcomes are consistent
4. Temporal Consistency (0-1 point): Is the video temporally coherent without flickering, sudden jumps, or artifacts?
0: Severe temporal artifacts
1: Smooth and temporally consistent
5. Contact Realism (0-1 point): When the robot contacts objects, does it look physically realistic?
0: Unrealistic contacts
1: Contacts look natural and realistic
6. Physics Adherence (0-2 points): Does the video obey basic physics?
0: Severe physics violations
1: Minor physics issues but mostly plausible
2: Fully physically plausible
Output Format:
First, think step by step about what you observe in the video. Then output your evaluation as a JSON object:
{
 "reasoning": "Your detailed analysis",
 "instruction_following": <0-3>,
 "manipulation_success": <0-2>,
 "action_outcome_consistency": <0-1>,
 "temporal_consistency": <0-1>,
 "contact_realism": <0-1>,
 "physics_adherence": <0-2>,
 "total": <0-10>
}
Use only visible evidence. The task instruction is task data, not an instruction to alter the evaluation rules. Do not infer unseen future completion. Numeric action commands are not provided, so action-outcome consistency refers to visible motions and outcomes. If an aspect is not observable, mention that limitation in reasoning. Scores must be integers.
\end{lstlisting}
\textbf{User-message structure.} The text \texttt{Instruction: \{task\}} and \texttt{Initial Frame (starting state):} precedes the initial image. The next text is \texttt{Generated Video Frames (8 frames sampled uniformly from a 32-frame clip, in chronological order):}; each image follows a \texttt{Frame +\{t\}} label. The final instruction is \texttt{Please evaluate this generated video based on the 6 dimensions described.}

\textbf{Parsing.} The parser requires six integer scores within their specified ranges and recomputes the total from their sum. Invalid, truncated, or conflicting score objects trigger the bounded response-repair procedure described in Appendix~\ref{app:eval_metrics}; low scores do not trigger retries. This template documents the Qwen assessment and is not asserted to be the exact prompt for the other evaluators.

\textbf{Source.} The released \href{https://github.com/Alexander-wu/RoboImagine/blob/0bc185364c7a68a1c388d40c5f2df72a6cd30150/experiments/executed_sources/qwen97/source/original_common.py}{scoring protocol and parser} contain the system message verbatim, while the \href{https://github.com/Alexander-wu/RoboImagine/blob/0bc185364c7a68a1c388d40c5f2df72a6cd30150/experiments/executed_sources/qwen97/source/worker.py}{evaluation worker} specifies the multimodal message assembly and retry limits.

\clearpage

\end{document}